\documentclass{IEEEoj}
\usepackage{cite}
\usepackage{amsmath,amssymb,amsfonts}
\usepackage{algorithmic,algorithm}
\usepackage{hyperref}
\usepackage{booktabs}
\usepackage{graphicx,color}
\usepackage{textcomp}
\def\BibTeX{{\rm B\kern-.05em{\sc i\kern-.025em b}\kern-.08em
    T\kern-.1667em\lower.7ex\hbox{E}\kern-.125emX}}
\AtBeginDocument{\definecolor{ojcolor}{cmyk}{0.93,0.59,0.15,0.02}}
\def\OJlogo{\vspace{-14pt}\includegraphics[height=28pt]{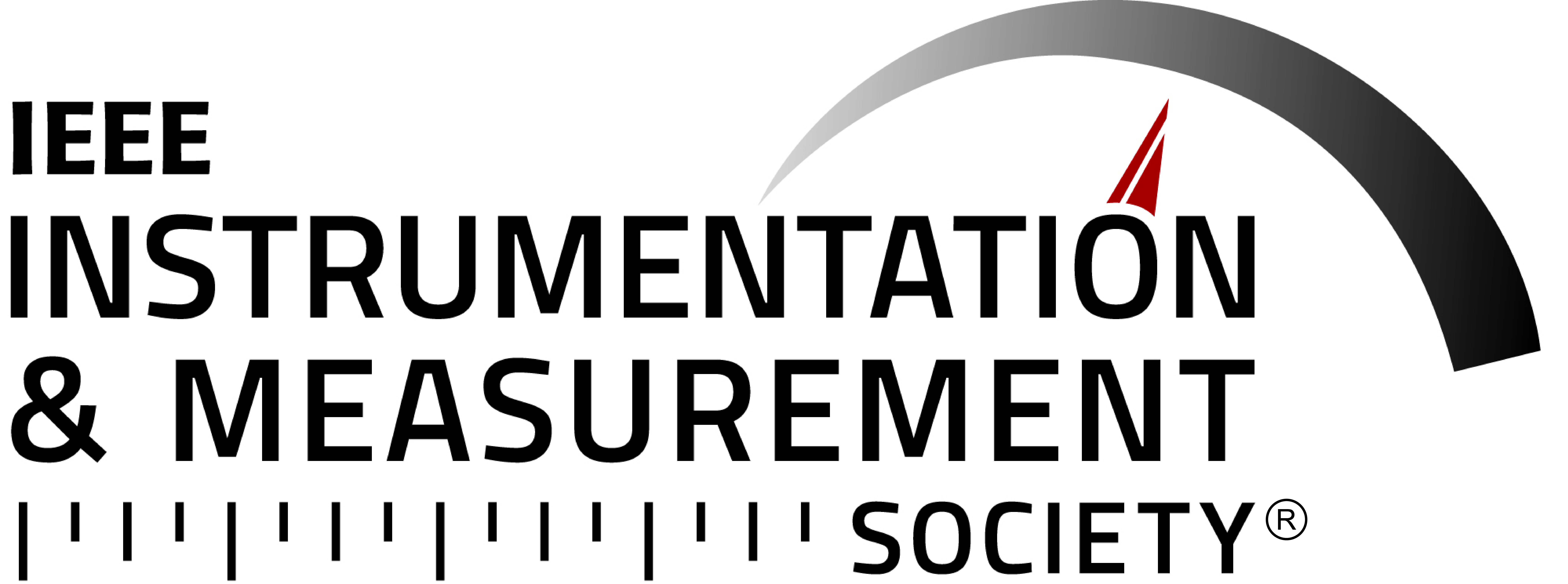}}
\begin{document}
\receiveddate{XX Month, XXXX}
\reviseddate{XX Month, XXXX}
\accepteddate{XX Month, XXXX}
\publisheddate{XX Month, XXXX}
\currentdate{XX Month, XXXX}
\doiinfo{OJIM.2022.1234567}

\title{A Computer Vision Based Approach for Stalking Detection Using a CNN-LSTM-MLP Hybrid Fusion Model}

\author{Murad Hasan$^{1}$, Shahriar Iqbal$^{1}$, Md. Billal Hossain Faisal$^{1}$, Md. Musnad Hossin Neloy$^{1}$, Md. Tonmoy Kabir$^{1}$, Md. Tanzim Reza $^{1}$, Md. Golam Rabiul Alam$^{1}$, Md Zia Uddin $^{2}$
}

\affil{Department of Computer Science and Engineering, BRAC University, Dhaka 1212, Bangladesh}
\affil{SINTEF Digital, Oslo, 0373, Norway}

\corresp{E-mail: murad.hasan@g.bracu.ac.bd, shahriar.iqbal@g.bracu.ac.bd, md.billal.hossain.faisal@g.bracu.ac.bd, md.musnad.hossin.neloy@g.bracu.ac.bd, md.tonmoy.kabir@g.bracu.ac.bd, tanzim.reza@bracu.ac.bd, rabiul.alam@bracu.ac.bd, zia.uddin@sintef.no}

\markboth{}{Hasan \textit{et al.}}

\begin{abstract}
Criminal and suspicious activity detection has become a popular research topic in recent years. The rapid growth of computer vision technologies has had a crucial impact on solving this issue. However, physical stalking detection is still a less explored area despite the evolution of modern technology. Nowadays, stalking in public places has become a common occurrence with women being the most affected. Stalking is a visible action that usually occurs before any criminal activity begins as the stalker begins to follow, loiter, and stare at the victim before committing any criminal activity such as assault, kidnapping, rape, and so on. Therefore, it has become a necessity to detect stalking as all of these criminal activities can be stopped in the first place through stalking detection. In this research, we propose a novel deep learning-based hybrid fusion model to detect potential stalkers from a single video with a minimal number of frames. We extract multiple relevant features, such as facial landmarks, head pose estimation, and relative distance, as numerical values from video frames. This data is fed into a multilayer perceptron (MLP) to perform a classification task between a stalking and a non-stalking scenario. Simultaneously, the video frames are fed into a combination of convolutional and LSTM models to extract the spatio-temporal features. We use a fusion of these numerical and spatio-temporal features to build a classifier to detect stalking incidents. Additionally, we introduce a dataset consisting of stalking and non-stalking videos gathered from various feature films and television series, which is also used to train the model. The experimental results show the efficiency and dynamism of our proposed stalker detection system, achieving 89.58\% testing accuracy with a significant improvement as compared to the state-of-the-art approaches.
\end{abstract}



\maketitle

\section{INTRODUCTION}
\IEEEPARstart{I}{n} the modern era, our lifestyle is constantly expanding and changing because of the tremendous strides that have been made in science and technology. The rapid advancement of digital technologies has created a plethora of new options for us to improve our lifestyle. By lowering the difficulty and the effort required to solve a task, we may apply our understanding of fundamental development in technological innovation to promote human well-being. To illustrate, a diverse range of problems and crimes are being resolved through the use of various automation technologies such as artificial intelligence, machine learning, deep learning, computer vision, and camera networking.

Violence against women is one of the most disturbing aspects of crime in today's world, with physical assault and cyberbullying being some of the exemplary forms. The form of violence is usually contingent upon the circumstances and the nature of the perpetrator's relationship with the victim. Several incidents of violence against women garnered widespread attention in recent times and were addressed via the use of information and communication technologies. However, a great number of different kinds of violence go unaddressed due to the lack of eyewitnesses. These violent activities usually begin with the victim being stalked by the criminal, which is a distinct type of behavior in which one individual harasses or monitors another by observing, following, spying, or looking at them. Other academics have defined stalking in a variety of ways in numerous philosophical research papers and publications. The researchers narrated the term stalking as the malicious, purposeful, and iterative pestering of another person, which is generally accompanied by genuine intimidation of harm against the victim \cite{r1}. As per medical experts, it's defined by intrusive actions (e.g., threatening, following, spying, getting unsolicited phone and email calls) that provide the victim with a false sense of security and make her believe she's in a dangerous situation that she has no control over \cite{r1}.  The definition of stalking in terms of criminology includes a wide variety of acts that are established by the precinct, local laws, and other types of systematic discretion. In criminology, stalking analysis has developed into a significant social issue. 

Stalking can be classified into two types: One is cyberstalking, and the other is street stalking, or in certain ways, physical stalking. Among these two forms, cyberstalking has already captured the majority of the public's attention and has sparked enormous public outrage in recent years. Additionally, many people are working diligently to put an end to cyberstalking. However, if we look closely, there has not been much systematic work on physical stalking detection. While information and communication technologies are advancing and improving across all spheres of social life, they have not kept pace with this cultural shift in instances of physical stalking. Physical stalking has a devastating effect on women, with the consequences often being very long-term and severe. Statistics from a variety of countries illustrate the severity of physical stalking. Bianca Fileborn examined 292 people across Australia in 2016 and discovered that 65.1\% of them were gazing at women on the street \cite{r2}. As a result, the victim's physical, mental, and social lives are severely impacted, as is her overall quality of life. Meanwhile, camera footage has been important in identifying a potential stalking situation. We can use computer vision, machine learning, and deep learning to solve the crime of physical stalking.

Nowadays, stalking women in public places is a widespread problem across the world. However, we are not sufficiently aware of the dangers associated with this widespread problem. A study says every woman and girl in Indianapolis, Indiana, USA, has encountered street harassment, particularly stalking, by unknown men in public \cite{r2}. As a result, many women and girls are forced to take a step back or are unable to walk freely on the street. Despite the pervasiveness of information and communication technology in many spheres of social life, physical stalking has not been a factor that has been put much thought into. Numerous academics, writers of computer science, and technologists have focused on detecting various human behaviors, but physical stalking detection has yet to gain the prominence it deserves among these behaviors. Though there is some work on collaborating with computer vision techniques or other algorithms, it should be increased since it is one of the first stages of every major crime. Furthermore, according to data from all around the world, stalking should be given top attention, with urgent steps taken to reduce the incidence of stalking.

\subsection{RESEARCH CONTRIBUTION}
In our research, we seek to identify various cases of stalking. To be more specific, we propose a feature fusion model combining CNN, LSTM, and MLP incorporating facial landmarks, head pose angles, relative distance, and numerical image frames to determine whether or not a stalking scenario is happening at a particular moment within a minimal number of frames. Our ultimate goal is to deploy our model in a surveillance system to detect suspicious physical stalking activity and classify it as stalking or non-stalking. This application can further help security enforcement authorities to identify and get evidence against any potential stalker. Furthermore, it will safeguard everyone on the streets by providing proof of possible stalking so that stalkers may be identified. The main contributions of this work are summarized below:
\begin{itemize}
    \item We propose a computer vision-based solution to detect stalking scenarios using deep learning techniques such as CNN, LSTM, and MLP. We propose a hybrid fusion model by utilizing these deep learning techniques that are able to classify stalking and non-stalking videos with a minimal number of frames.

    \item We demonstrate how facial landmarks, head position angles, and relative distance can be included in activity recognition tasks when facial movements are crucial. Also, utilizing these features considerably improves the accuracy of the model and provides dynamism to solve the problem.

    \item We also present a video dataset consisting of both stalking and non-stalking videos. It is the first publicly available dataset that includes single videos of stalking and non-stalking scenarios. These videos are capable of demonstrating the differences between these two scenarios without the need for multiple appearances.

    \item Experimental results on our new dataset show that our proposed method is able to distinguish between stalking and non-stalking behaviors from a single appearance-based video with a minimal number of frames, which is a significant achievement as compared to the state-of-the-art approaches.

\end{itemize}
The rest of this paper is structured as follows: The literature overview of previous work is presented in Section~\ref{RELATED WORK}. The materials and methods used in the study are described in Section~\ref{MATERIAL AND METHODS}. Section~\ref{EXPERIMENTS AND RESULTS} includes the experiments and results, where a complete analysis of the outcome of each model is shown. Finally, the research conclusions and future work are presented in Section~\ref{CONCLUSION}.

\section{RELATED WORK}
\label{RELATED WORK}
Regardless of the fact that security is one of the most important aspects of an individual's everyday existence, very few studies have been conducted on identifying stalking. Stalking is viewed as a dangerous offense in the modern world due to the direct and imperceptible effects it has on our society. In our very neighboring country, India, a survey on street harassment of women was conducted \cite{r3}. This report's outcome was alarming as almost 67\% of those polled said they had been stalked on the street. To solve this problem, the researchers have primarily focused on examining suspicious behaviors exhibited by individuals. Many of these researchers have highlighted the use of human motion, activity, and actions as integral components in the authors' models. Only one of these studies has concentrated on the challenging task of detecting stalking behaviors. This section aims to represent the many efforts that have been conducted in the field of detecting suspicious human activity and determining stalking incidents.

Dandamudi et al. (2020) \cite{r4} primarily discussed a CNN-based model for drone image technology. The authors proposed a model in which each person is detected within a frame, and pose estimation is used to identify humans using key points. Additionally, this model recognizes human positions on the same scale. The significance of this model is that it enables a more precise classification of human actions. However, no head and eye movement was included there, whereas stalking situations include such movements; therefore, this model will not be able to identify crimes like stalking that involve any obvious head movements on the part of the victim.

Bhattacherjee and Somashekhar (2017) \cite{r5} developed a flying device that resembles a bird and is capable of continuously monitoring any suspicious activity that occurs in its vicinity. It is capable of locating suspects or any human being with the bare minimum of manual human effort. This intelligent device is also responsible for transmitting a signal to the nearest police station containing the suspect's location and for tracking the suspect's movement. This device is schematically similar to a bird that is following an aerodynamic model and is equipped with a 360-degree rotational camera. Additionally, it makes use of an Arduino Uno processor and a GPS tracker worn on the body to navigate to the suspected object's location. This model makes use of a MatLab image comparison algorithm (SPIHT ALGORITHM) to reconcile the latest image with the reference image. However, it will not be economical or efficient because the gadget resembles a bird that must constantly fly to detect anything odd.

Sidhu and Sharad (2016) \cite{r6} considered two distinct application areas in order to ensure confidentiality and security in a variety of locations. One is in an office environment where CCTV cameras begin recording when the likelihood of a "critical situation" occurring is high (triggered on). Second, in public locations, they conduct a preliminary examination of critical situations such as bullying, harassment, and assault. The researchers used video clips and voice data in their proposed model to accomplish their goal. Although recording voice data in public places is difficult, their primary goal was to achieve their objective in office environments.


Pang et al. (2014) \cite{r8} and He et al. (2011) \cite{r9} conducted studies on the same sort of human behavioral activities. Their primary objective was to recognize behaviors such as walking, jogging, sprinting, hand waving, leaping, skipping, punching, and kicking, among many others. Pang et al. \cite{r8}, in their research, indicated that by analyzing the location of bodily joints, normal and aggressive behavior can be distinguished. For the study of bodily joints, they used the Microsoft Kinect sensor and the Cartesian coordinate formula in their model. In contrast, He et al. \cite{r9} identified human activities using the double-layer Bag-of-Words model and the Latent Dirichlet Allocation model. The authors \cite{r9} worked only on a single action video clip. However, the authors of these two papers encountered several challenges. For example, Pang et al. discovered that when human interaction is low, it is difficult to distinguish normal actions such as jumping and walking because there is no need for two or multiple persons physical interaction, and He et al. encountered difficulties identifying multiple actions from the same video due to a lack of large and challenging datasets.

Tiwari et al. (2015) \cite{r10} claimed that distinguishing suspicious face detection could also be a game changer. Eye movement, according to cognitive-visuomotor theory, is a very rigid phenomenon that characterizes a suspicious scenario. A person's thought at a particular moment can be deduced from his or her glancing, as proposed by the "Eye-Mind Hypothesis." The non-linear entropy of a criminal's eyes is significantly greater than that of a normal person. Additionally, rapid eye movement is another indicator of suspicious patterns. Nominating eye movement is an effective technique for setting the ambit stalkers group apart. Nevertheless, it is uncommon to get data where exact eye movements may be recognized. Most of the time, the video data we receive is not sufficiently detailed to allow us to see the iris or eye movement.

Shakya et al. (2016) \cite{r11} proposed another technique for detecting human behavior using facial expression analysis. At first, video data is converted into an image sequence. Afterward, the faces, noses, eyes, mouths, and upper bodies of people are identified using a color-filtered image algorithm. There are a number of telltale signs, such as the suspect's wrathful lips, the corners of his mouth showing anguish, and his crimson complexion, which shows either intoxication or anger. Additionally, a PCI algorithm was used to extract features, which calculate the coefficient, latent, and score. Additionally, the researchers used Euclidian distance calculations to predict facial expressions. All algorithms were implemented using the Viola-Jones AdaBoost technique. The HMM, Bayesian, or Kalman methods were used to track a moving person.

Bisagno et al. (2018) \cite{r12} insisted that calculating crowd flux could be an extremely effective tool for behavior detection. An advantageous use of resolution, which includes paying special attention to crowd scenes, can enhance surveillance's dynamic nature. Providing the same resolution throughout the surveillance area can occasionally result in a loss of resolution. Thus, using a constant population to determine whether or not a space is crowded is a very noble idea. Additional resolution in the dense zone can help ensure the model's accuracy. Dividing the entire area into local and global coverage zones is an excellent idea. This strategy ensures a minimum resolution in each segment of the area, with an emphasis on the crowded zone, resulting in a highly productive, cost-effective, and practical system. Another method of determining if a location is crowded is to analyze a data set. Additional surveillance of those alarming areas may make the approach more effective.

Apart from the numerous efforts made to detect suspicious behavior, only one research study has focused on detecting stalker patterns. Liu et al. (2019) \cite{r13} identified a new stalker pattern through a noble parameter dubbed "spatiotemporal co-appearance". This means that the same pair of people have been accused of stalking in different time periods. Neo-face detection is successfully used to identify the same person in multiple frames. Despite the fact that this paper accomplished a great deal, this system will remain focused on a single point. For instance, stalking cases will be detected when friends, family members, and couples walk side by side. Too many false-positive cases could erode the system's usefulness in practice. Furthermore, their system identifies a potential stalking scenario based on different frames of a different scenario. The system requires a lot of frames from several scenarios involving the same person. However, the same person cannot always be detected in different scenarios, despite being in the original stalking situation. Hence, this paper also has its own limitations, which are not too small to ignore.

The preceding research papers have primarily focused on examining suspicious behaviors exhibited by individuals. Many of these papers have highlighted the use of human motion, activity, and actions as integral components in the authors' models. One of these studies has concentrated on the challenging task of detecting stalking behaviors, but they have encountered difficulties in obtaining the necessary ideal data. Authors have employed a method that involves analyzing multiple videos to identify stalking, which can be costly and reliant on a wide range of image frames captured under various scenarios. Multiple appearances of the same person, no additional feature extraction without facial recognition, and the requirement for person re-identification made this effort expensive. These constraints influenced our work to mitigate such issues and detect stalking from a single video.

\section{MATERIAL AND METHODS}
\label{MATERIAL AND METHODS}
\begin{figure}[htp]
    \centering
    \includegraphics[width=3.4in]{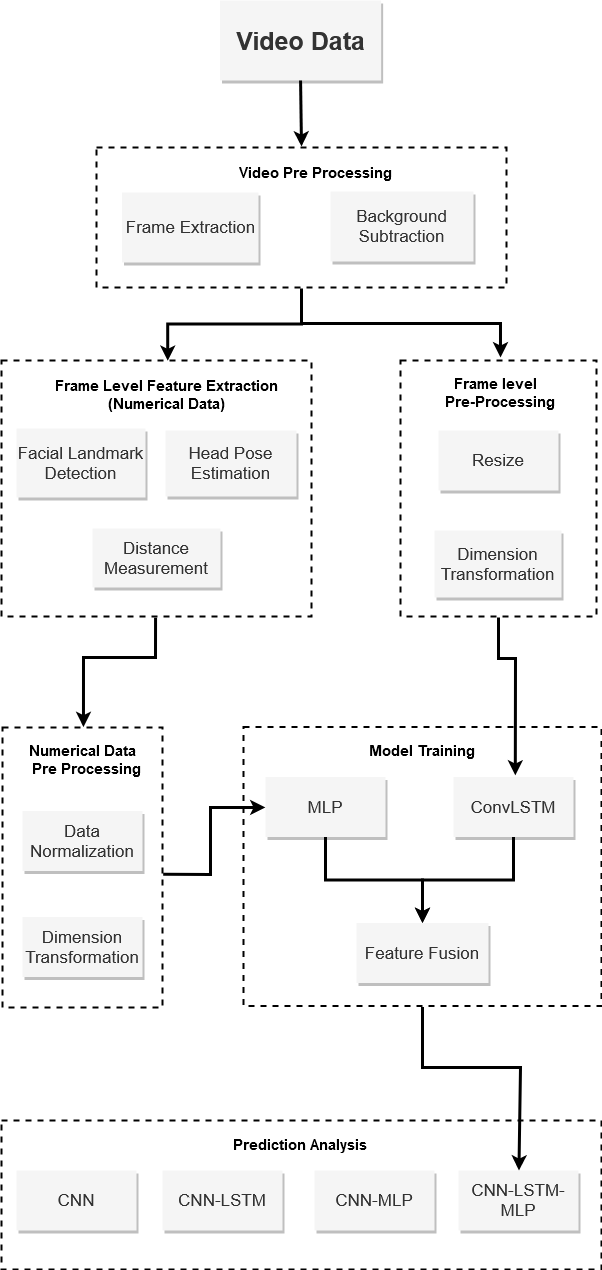}
    \caption{ Top-level view of the proposed model. }
    \label{fig1}
\end{figure}

A workflow of our research has been demonstrated in Fig.~\ref{fig1}. The proposed system model consists of a few steps. Initially, video image frames are extracted, and the backgrounds of the images are removed. Facial landmarks, facial angles, and relative distance are obtained as numerical values using these image frames. The numerical values are then pre-processed using a data normalization technique, and the dimensions are transformed before being fed into the Multilayer Perceptron (MLP) model. Concurrently, the video images are also resized and transformed before being fed into a convolutional neural network (CNN) and long short-term memory (LSTM) model combination. In a feature fusion approach, the outputs of both models are concatenated. Finally, the feature fusion result completes the classification task by predicting whether a video entails stalking or not. For comparison purposes, we also employ other methods, whereas our proposed method outperforms others.

\label{sec:sample2}

\subsection{DATASET DESCRIPTION}
Stalking detection in the field of computer vision is a relatively unexplored area, resulting in a scarcity of publicly available real-world datasets in this domain. Due to the limitations of the dataset, we decided to create our own unique dataset to evaluate our model. For our research, we require video footage depicting one person stalking another, which is a particularly challenging type of content to obtain. Our initial data collection efforts include various sources, with YouTube being a major contributor. However, a significant obstacle we encounter is that the stalkers' faces are often obscured in most of the YouTube videos, making it difficult to gather an adequate amount of relevant video data. Eventually, we managed to accumulate a substantial portion of our dataset from feature films and television series.

In this dataset, there are a total of 238 videos included. The dataset comprises 117 videos belonging to the stalking class and 121 videos belonging to the non-stalking class, amounting to a total size of 2.1 GB. Each video has a runtime ranging from around 3 to 8 seconds. The average size of each video is approximately 15 megabytes. The quality of the retrieved videos is high, as they are sourced from feature films and television series. Later, for ease of computation, the videos are transformed into the same dimension. In order to strengthen the dataset, we individually annotate each video. Out of the initial set of 295 gathered videos, a total of 57 videos were eliminated due to factors such as obscurity, individuals wearing caps, masks, and other similar reasons. The final dataset is determined by taking the majority of votes for video-level labels as well as the clarity of distinguishment of two scenarios from five different annotators.

\label{sec:sample3}
\subsection{FRAME EXTRACTION}
According to the previous study on stalking, it occurs over a longer period of time \cite{r1}. Multiple behaviors, including loitering, following, observing, harassing, and interfering with personal property, are the first steps in a stalking scenario. However, the actual stalking, which includes gazing and interfering with personal property, only lasts for a short period of time. As a result, our dataset contains trimmed videos ranging from 3 to 8 seconds in length where stalking occurs. Therefore, we extract frames from the videos. Only 5 frames from each video have been extracted for these reasons. From 1 second, approximately 1 frame has been taken. For purposes of comparison, we examine more than 5 frames from each video that have minor changes from one frame to the next. Thus, a total of 1190 frames are retrieved from the full dataset by taking 5 frames from each video.

\label{sec:sample3}
\subsection{BACKGROUND SUBTRACTION}
\begin{figure}[!tbp]
  \centering
  \begin{minipage}[b]{0.5\textwidth}
    \includegraphics[width=3.4in]{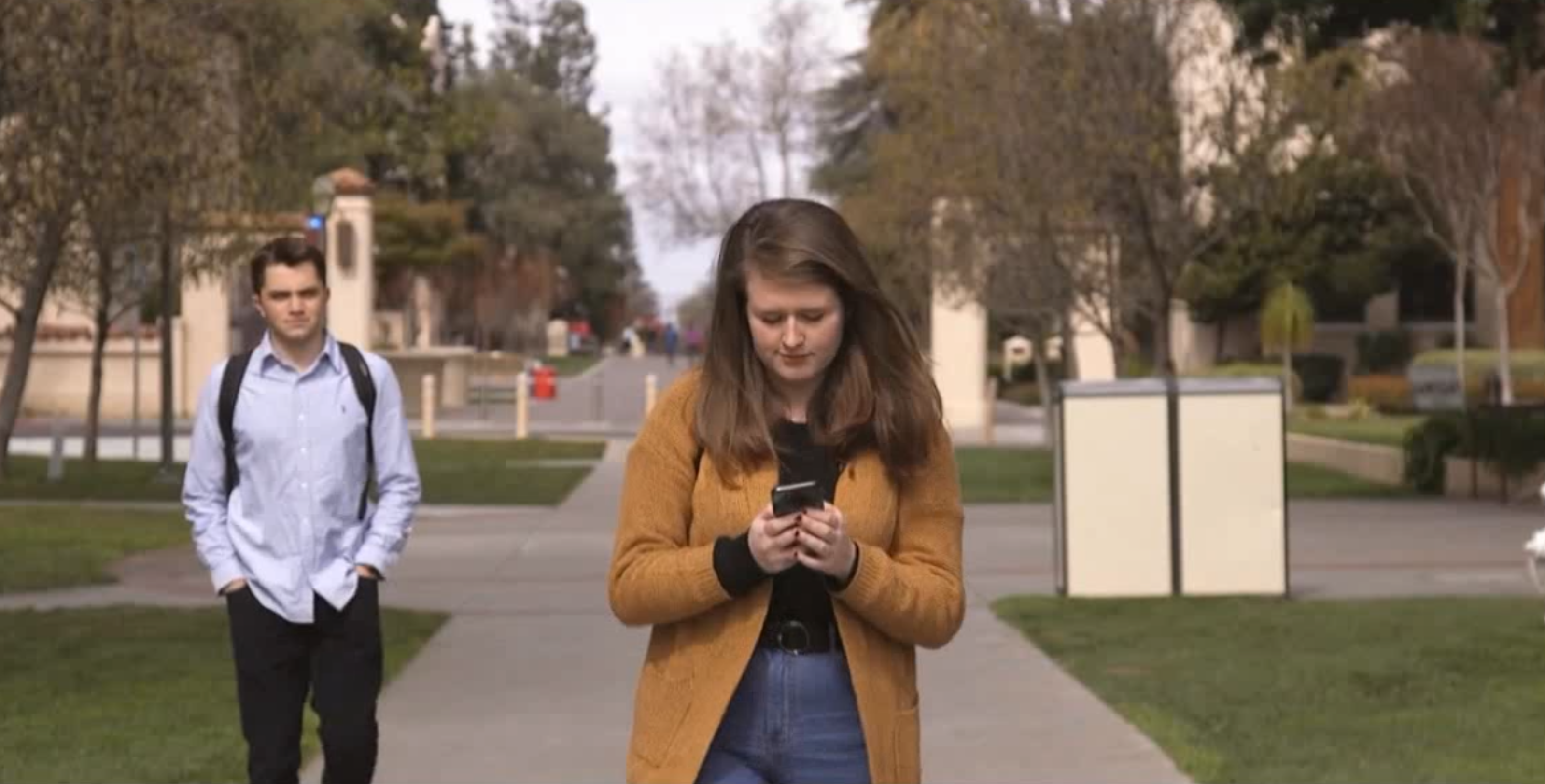}
\centering(a) Video image frame with background.
  \end{minipage}

  \hfill
  \begin{minipage}[b]{0.5\textwidth}
    \includegraphics[width=3.4in]{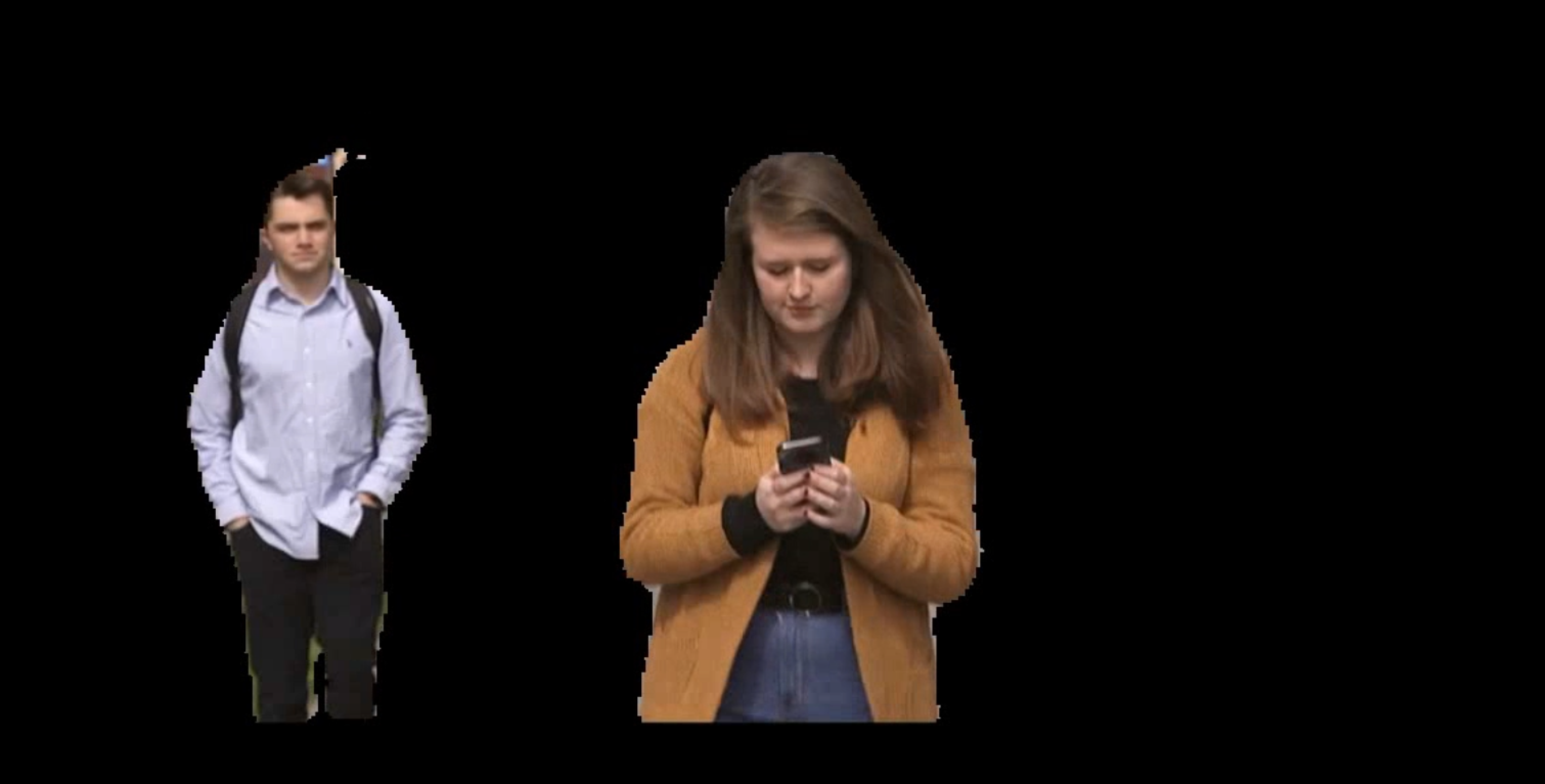}
\centering (b) Video image frame without background.
  \end{minipage}
    \caption{ A comparison is made between video frame with backgrounds and video frame without backgrounds.}
    \label{fig2}
\end{figure}

As deep learning technology has developed rapidly in recent years, computer vision researchers have achieved great progress in object identification. In our model, we only need the humans from the image frames. Other objects may create noise and complexity for the model. Hence, subtracting the background along with other objects except humans is crucial for our model. Therefore, we need to use models that can efficiently detect objects, subtract the background, and keep only humans. There are two widely popular models now used in object detection techniques, such as Mask R-CNN \cite{r14} and YOLO \cite{r15}. Yolo's method is faster, but it has a few drawbacks, such as the fact that it can produce false positives in the background area. It also has a low average precision compared to the Mask R-CNN, making it a poor performer overall \cite{r16}. In addition to this, Yolo has a lower recall rate and a higher precision error rate. Furthermore, despite Yolo's high detection rate, there is a noticeable fall in accuracy when it comes to recognition tasks. In paper \cite{r17}, using 2049 images of license plates, Yolo achieved a recognition rate of 78\%, whereas in paper \cite{r18}, Mask R-CNN achieved a precision rate of 97.31\%, even in overlapping object detection, indicating a notable improvement of accuracy. Therefore, in our study, we employ Mask R-CNN to do background subtraction from video image frames. The PixelLib library \cite{r19} is used for the purpose of background subtraction. PixelLib is a Python programming language library renowned for its proficiency in addressing segmentation challenges through the utilization of a powerful Mask R-CNN model. This library is not limited to image segmentation but also extends its capabilities to video segmentation. A sample output of background subtraction is shown in Fig.~\ref{fig2}.

\label{sec:sample3}
\subsection{FACIAL LANDMARK DETECTION}
In our system, face component analysis is very crucial for identifying real instances of stalking. Accurate face landmarks and facial feature detection are crucial processes that influence subsequent face-centered activities \cite{r20}. Facial landmarks are a very efficient method for assessing facial points. It plays a crucial role in several image processing and computer vision applications. Thus, face landmark characteristics are important features in our fusion model.

The definition of a facial landmark is the identification and localization of distinct facial points. Eye corners, the tip of the nose, the corners of the nostrils, the corners of the mouth, the endpoints of the eyebrow arcs, earlobes, the nasion, the chin, etc., are common landmarks. Face-related activities, including gaze detection, expression comprehension, face identification, face tracking, and face registration, are all influenced by these fundamental processes \cite{r21}. Similar to \cite{r22}, the cascade of regressors, learning each regressor in cascade, and tree-based regressor are utilized in this work to detect facial landmarks. To start with the cascade regressors, let us say we have an image of a face with p facial landmarks. We can represent the coordinates of these landmarks as a vector  $ v=  (A_1^T , A_2^T ,..., A_p^T ) ^ T  $ , where $A_{i}$ is the 2D coordinates of the I-th landmark.

 We can use a cascade of regressors to estimate the location of the facial landmarks. Each regressor $R^t$ takes an image and a current estimate of the facial landmarks $V^{(t)}$ as input, and outputs an update vector $V^{(t)}$. The update vector is then added to the current estimate to improve it:

\begin{equation}
    V^{(t+1)} = V^{(t)} + R_{t} (I, V^ {(t)})
\label{p1}
\end{equation}

The key point is that the regressor $R_t$ makes its predictions based on features such as pixel intensity values that are computed from the image and indexed relative to the current estimate of the facial landmarks $V^{(t)}$. This introduces some form of geometric invariance into the process, and as the cascade proceeds, we can be more certain that the regressor is indexing a precise semantic location on the face. The cascade of regressors can be used to estimate the location of facial landmarks in a variety of conditions, including different poses, lighting, and occlusions.

We employ the gradient tree boosting technique with a sum of square error loss to train each $R_T$. Considering that we have training data $(I_{1} , V_{1}) ,..., (I_{n},V_{n})$, Where each $I_i$ is a face picture and $V_i$  is its form vector. In order to learn the first regression function $R_0$ in the cascade, we construct triplets from our training data comprising a face picture, an initial shape estimate, and the target update step, that is,  $(I_{\pi i},  V_i^{(0)},  \Delta V_i^{(0)})$ where,

\begin{equation}
    \pi_i  \  \epsilon \  ({1,...,n})
\label{p2}
\end{equation}

\vspace{2mm}
\begin{equation}
 V_i^{(0)} \epsilon \  \{ V_i,...,V_n\}/V_{\pi i}  
 \label{p3}
\end{equation}

\vspace{2mm}
\begin{equation}
 \Delta V_i^{(0)} = V_{\pi i} - V_i^{(0)}  
 \label{p4}
\end{equation}

Where $i = 1,..., N$. The sum of these triplets is set to $N = nM$, where $M$ is the number of initializations applied to each picture $I_i$. An image's initial shape estimates are each uniformly sampled $ \{V_i,...,V_n\} $  without replacement. With the use of gradient tree boosting and a sum of square error loss, we are able to extract the regression function R0 from this data. Then, by setting (with $t = 0$), the set of training triplets is updated to give the training data $ (I_{\pi i}, V_{i}^{(0)}, \Delta V_{i}^{(0)} )$ , for the following regressor $R_1$.

\begin{equation}
    V_i^{(t+1)} = V_i^{(t)} + R_t \ (I_{\pi i}, V_i^{(t)} )
    \label{p5}
\end{equation}

\vspace{2mm}
\begin{equation}
 \Delta V_i^{(t+1)} = V_{\pi i} - V_i^{(t+1)} 
 \label{p6}
\end{equation}

This method is repeated until a set of $T$ regressors, $R_{0}, R_{1},...,R_{T-1}$, are learned and integrated to provide a degree of accuracy that is considered sufficient.

\begin{figure}[!tbp]
  \centering
  \begin{minipage}[b]{0.5\textwidth}
    \includegraphics[width=3.5in]{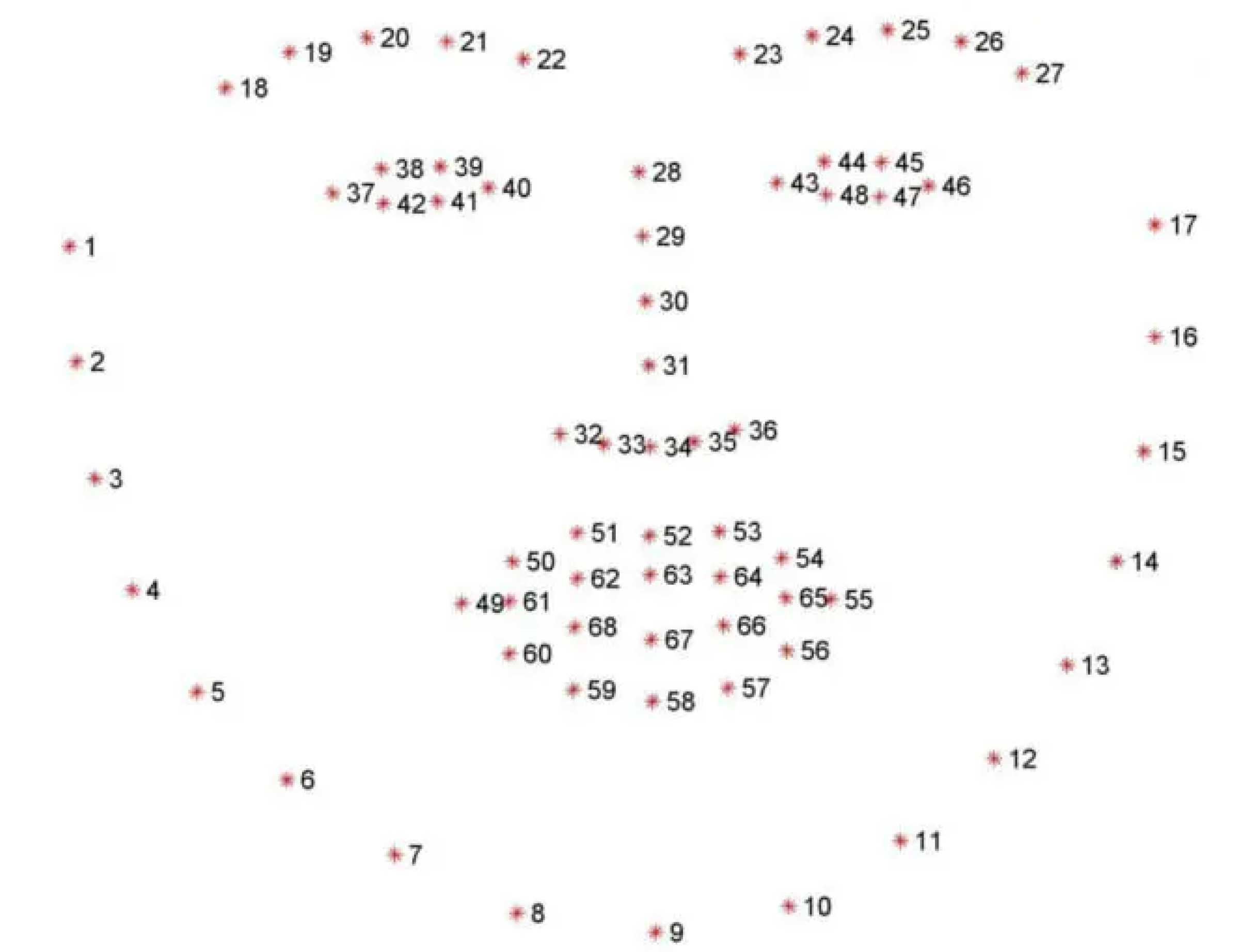}
\centering(a) All 68 facial landmark points.
  \end{minipage}

\vspace{5mm}

  \hfill
  \begin{minipage}[b]{0.5\textwidth}
    \includegraphics[width=3.5in]{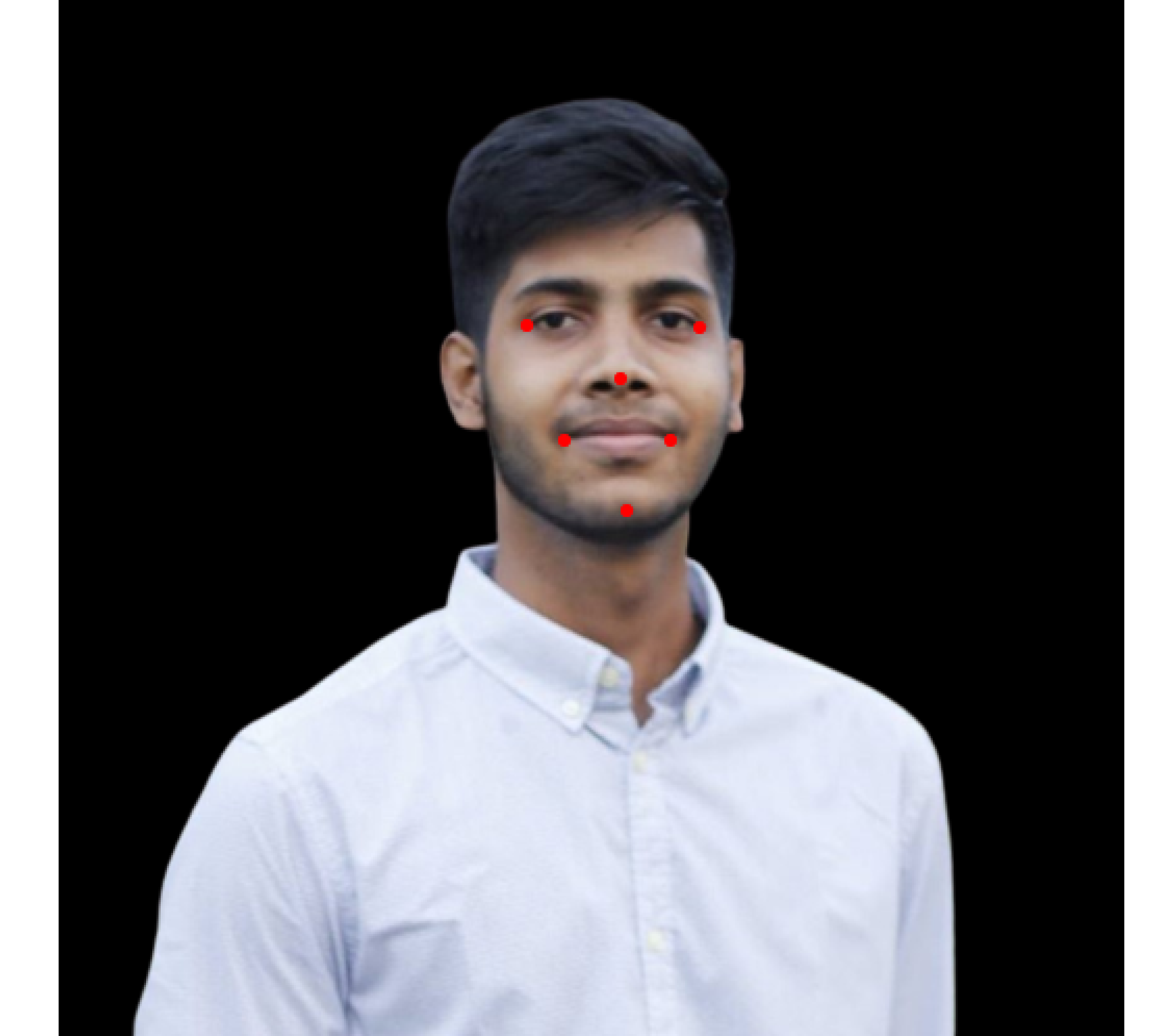}
\centering (b) Selected 6 points from 68 points.
  \end{minipage}
    \caption{ An illustration of facial landmark points (a) and their corresponding output after the selection of a limited number of points (b).}
    \label{fig3}
\end{figure}

The process of facial landmarks detection also follows the tree-based regressor which has shape invariant split tests, choosing the node splits and feature selection method as specified in Kazemi et al’s paper \cite{r22}. All these methods are already integrated into the dlib library \cite{r23} which provides a facial landmark detector that is implemented on the basis of Kazemi et al \cite{r22}. In this study, we also utilize a pre-trained facial landmark detector that is incorporated into the dlib library. Dlib’s 68-point facial landmark detector tends to be the most popular facial landmark detector in the computer vision field due to the speed and reliability of the dlib library. It includes a pre-trained facial landmark detector that can detect 68 points on a face. These points are identified from the iBUG300-W dataset, which is a large dataset of facial images with ground truth annotations for the location of 68 facial landmarks \cite{r23}. The pre-trained facial landmark detector inside the dlib library is used to estimate the location of 68 (x, y)-coordinates that map to facial structures on the face. The dlib facial landmark detector works by first detecting faces in an image using a HOG-based face detector. Once a face is detected, the landmark detector uses a regression model to estimate the location of the 68 landmarks on the face. The regression model is trained on the iBUG300-W dataset, so it is able to generalize faces in a variety of poses and lighting conditions. The indexes of the 68 coordinates can be visualized in part (a) of Fig. \ref{fig3}.

\begin{figure*}
  \centering
  \begin{tabular}{c c c c}
  \includegraphics[width=3.35in]{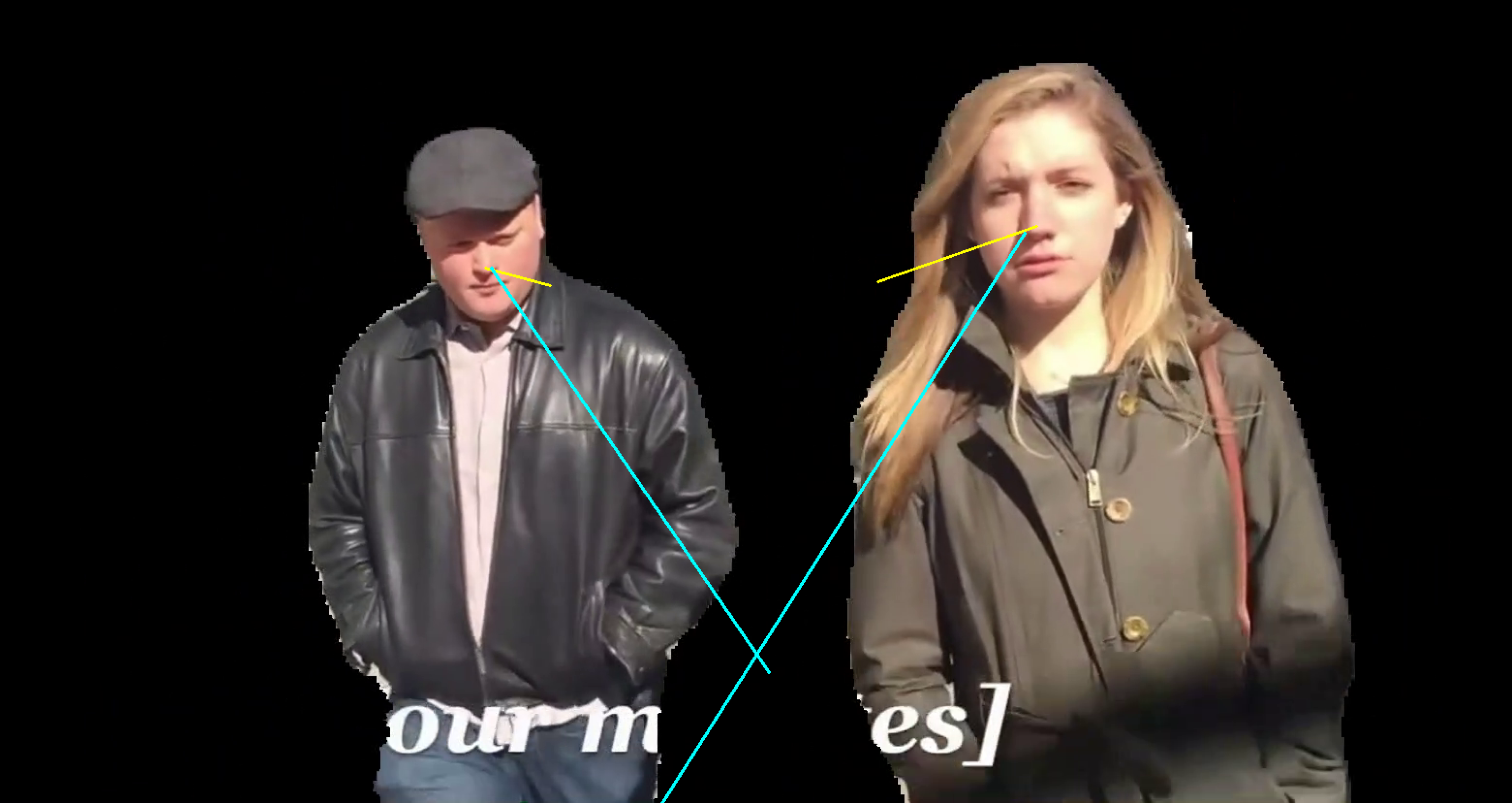} &
  \includegraphics[width=3.35in]{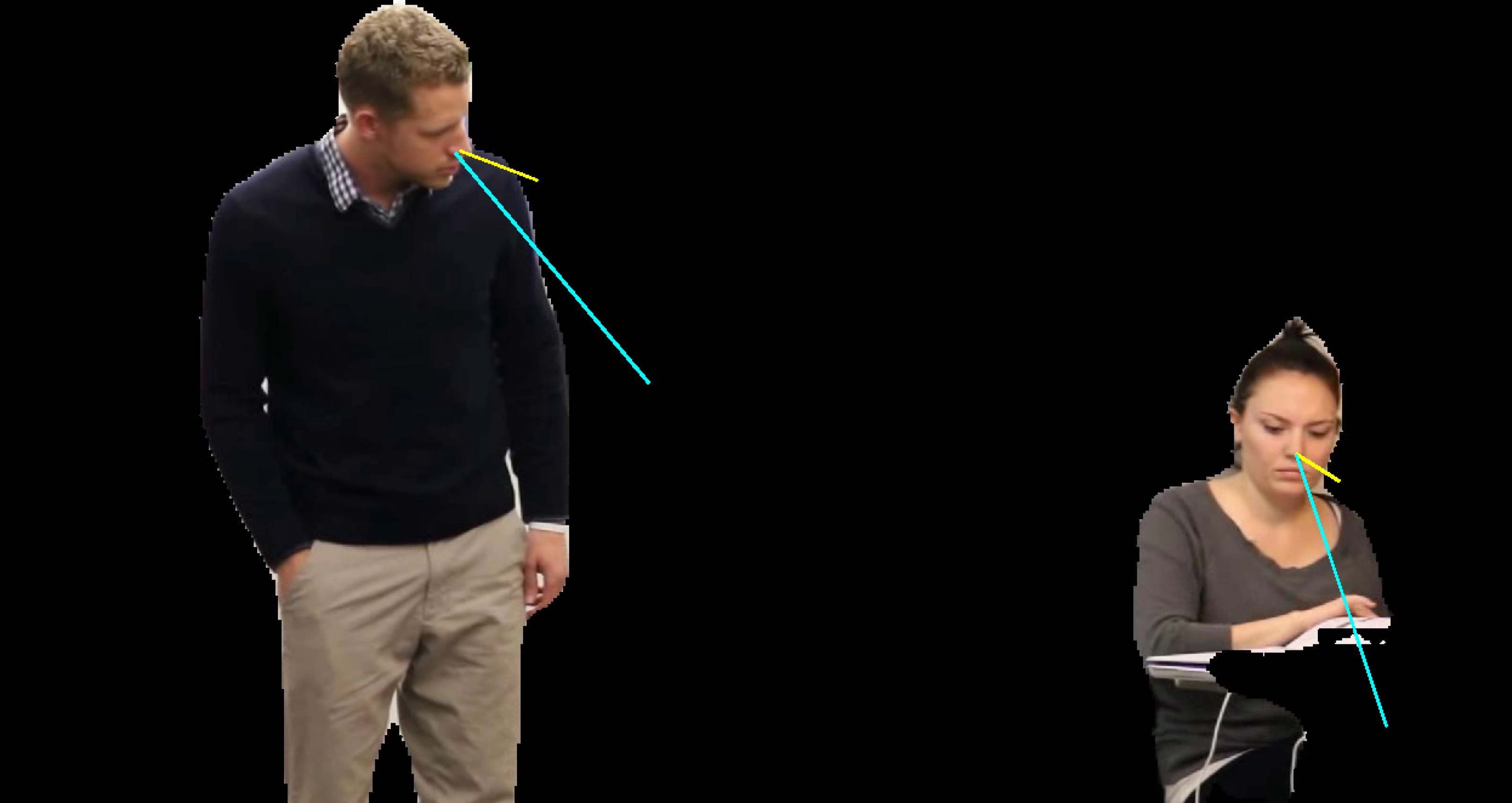} \\
  (a)~Stalking video frame.	 & (b)~Stalking video frame. &
  \end{tabular}
\end{figure*}

\begin{figure*}
  \centering
  \begin{tabular}{c c c c}
  \includegraphics[width=3.35in]{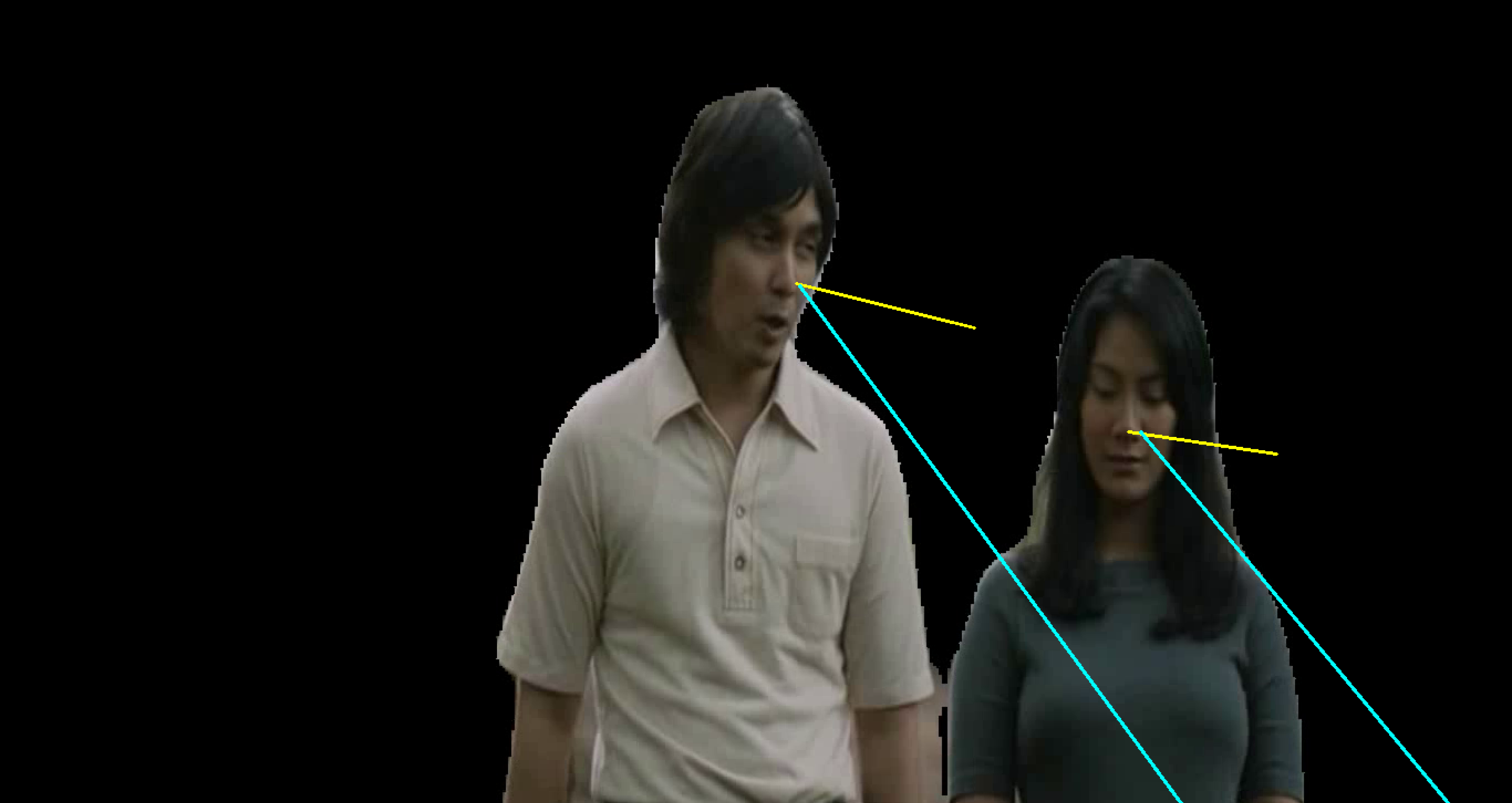} &
  \includegraphics[width=3.35in]{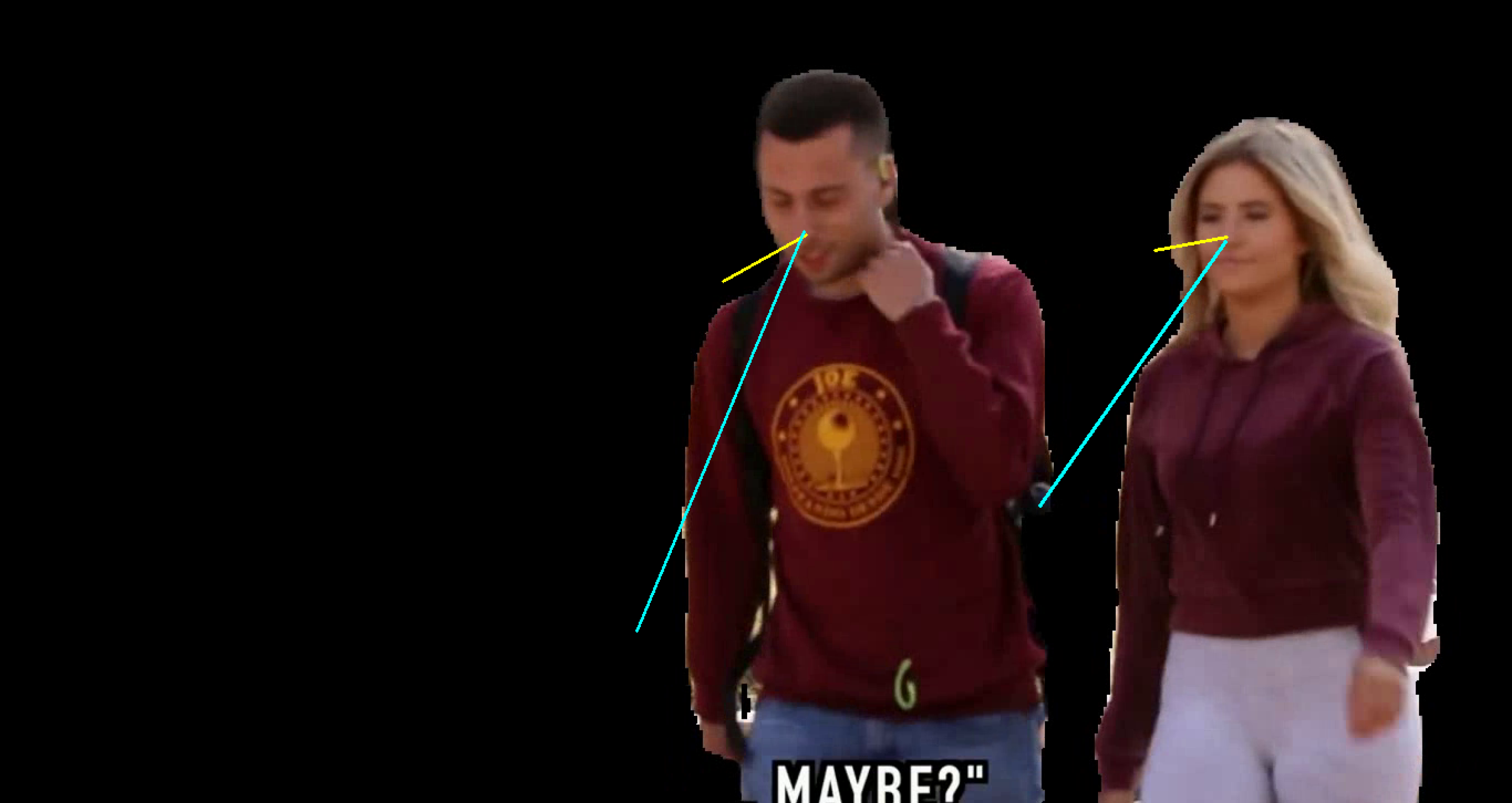} \\
  (c)~Non-stalking video frame. & (d)~Non-stalking video frame. & 
  \end{tabular}
  \caption{The output of head pose angles demonstrated by drawing lines on images.}
  \label{fig4}
\end{figure*}

The 68 points are allocated to six distinct facial components. Face forms range from 1 to 17, left and right eyebrows from 18 to 27, left and right eyes from 37 to 48, etc. The range for the nose is between 28 to 36, while the range for the mouth is between 49 and 68. Following the successful detection of the landmarks that correspond to our requirement, we choose some crucial points which are 34, 9, 37, 46, 49, and 55 for the nose tip, the chin, the left eye left corner, the right eye right corner, the left mouth corner and the mouth corner respectively. These six points are obtained using x and y coordinates. As a result, each person has 12 coordinates that are denoted by numerical values. Consequently, from each frame with a victim and a stalker, we obtain a total of 24 numerical values from facial landmarks (12*2 numerical values for two people). The six points that have been chosen are shown in part (b) of Fig.~\ref{fig3}.

\label{sec:sample3}
\subsection{HEAD POSE ESTIMATION}
In recent years, the detection and monitoring of head movements have been a very active field of study that significantly contributes to computer vision. Detecting angles of the head direction is used in a variety of disciplines, such as human behavior analysis, human-computer interaction (HCI), driver assistance systems, and pose-invariant face recognition \cite{r24}. Head movements play a crucial role in conveying a substantial amount of information during interpersonal communication. For instance, individuals tend to orient their faces toward each other while engaged in conversations. Additionally, nodding is commonly employed to indicate agreement or comprehension. Furthermore, when individuals shift their attention and become aware of something in close proximity that captures their interest, they typically rotate their heads in that direction. These are just a few examples of the diverse range of information that may be conveyed by head movements.

In short, head pose estimation refers to the process of identifying the precise position and orientation of a human head within an image. The determination of the Euler angles for yaw, pitch, and roll of the head has great importance \cite{r24}. In this context, Yaw refers to the angular displacement of a human head in the horizontal plane, namely the rotation around the Y-axis. Pitch, on the other hand, is generated by the vertical movement of the head, corresponding to the rotation around the X-axis. Lastly, roll denotes the inclination angle of the head, resulting from rotation around the Z-axis.

In our research, we utilize the six landmark points that were previously retrieved using the dlib library in order to compute the angles of the head pose \cite{r25}. The analysis suggests that roll angle, which represents rotation along the Z-axis, cannot produce any significant results when examined using 2D images. Therefore, we decided to remove the roll angle and only incorporate two fundamental components derived from head pose estimation, namely yaw and pitch. Two numerical values are obtained for each person, resulting in a total of four numerical values obtained for two individuals, one being the stalker and the other being the victim. Several examples of output lines produced at different angles are depicted in Fig.~\ref{fig4}.

\label{sec:sample3}
\subsection{RELATIVE DISTANCE MEASUREMENT}
Several studies \cite{r1,r26,r13} have explored how the relative distance between stalker and victim plays a role in understanding the situation. Measuring their relative distance is essential for determining whether a scenario defines stalking. For example, if two people are familiar with each other, they will usually stay close to each other, but if they are not, the scenario changes. In non-stalking incidents, two people are observed engaging in a variety of activities, such as talking, laughing, or staring at each other. However, the situation is different in the case of the opposite events, where they keep a distance since they do not know each other or try to avoid direct gaze. Therefore, measuring relative distance becomes necessary for our study. We measure the facial distance between two people to avoid complexity. Due to the fact that the facial landmarks have already been obtained, it becomes simpler to calculate the facial distance. We select the nose tip (point 32) from facial landmarks because it is located in the center of the face. The Euclidian distance method is used, utilizing the x and y coordinates of the nose tips of two people. The equation is defined as follows:

\begin{equation}
     d = \sqrt {({ V_{x} - S_{x} })^2  + ({ V_{y} - S_{y} })^2}
     \label{p7}
\end{equation}

where, $d$ represents the resulting distance, $V_x$ represents the victim’s nose tip’s x coordinate, $V_y$ represents the victim’s nose tip’s y coordinate, $S_x$ represents the stalker’s nose tip’s x coordinate, and $S_y$ represents the stalker’s nose tip’s y coordinate.

\vspace{5mm}
\begin{figure}[!tbp]
  \centering
  \begin{minipage}[b]{0.5\textwidth}
    \includegraphics[width=3.4in]{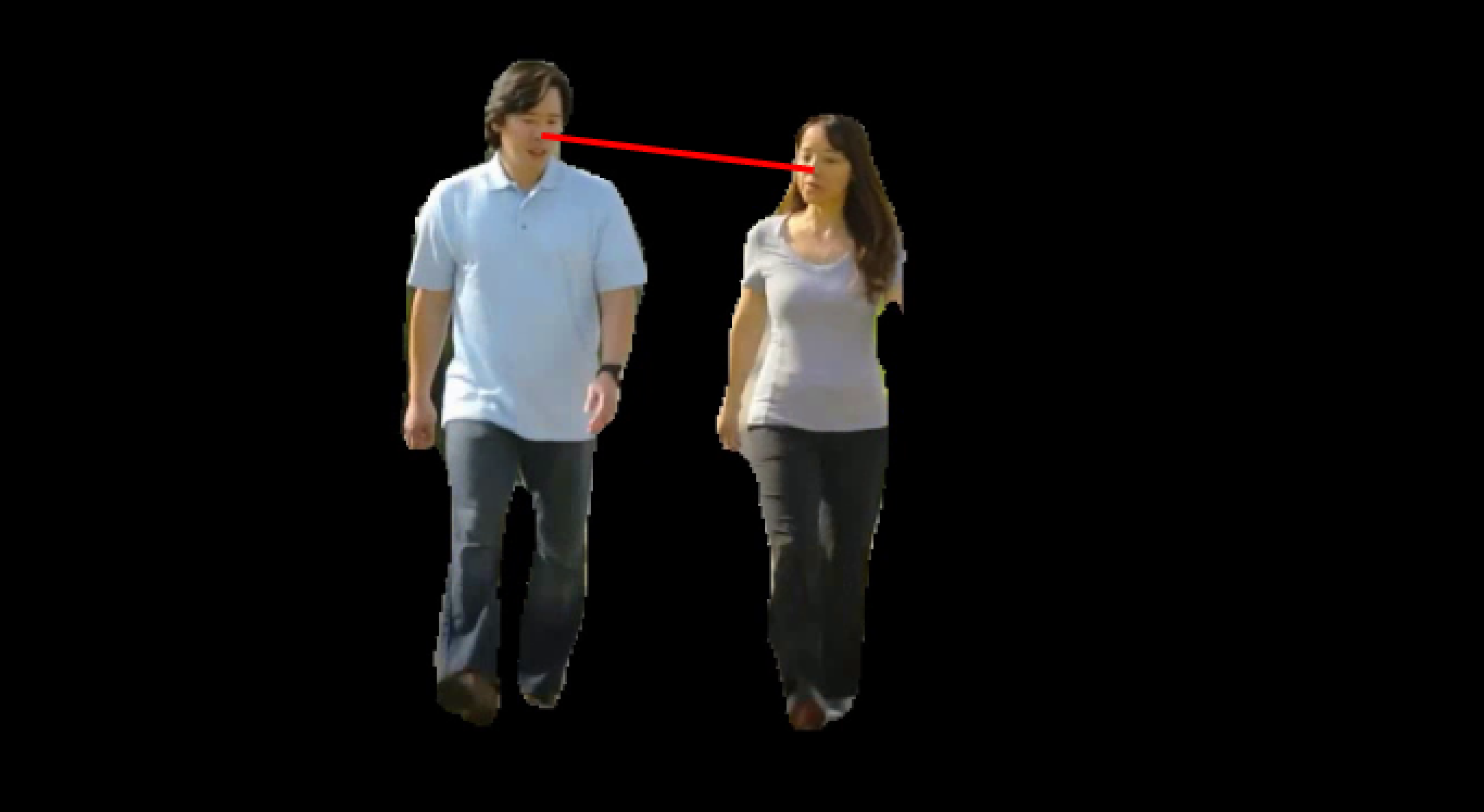}
    \centering(a) Non-stalking video frame.
  \end{minipage}

\vspace{5mm}
  \hfill

  \begin{minipage}[b]{0.5\textwidth}
    \includegraphics[width=3.4in]{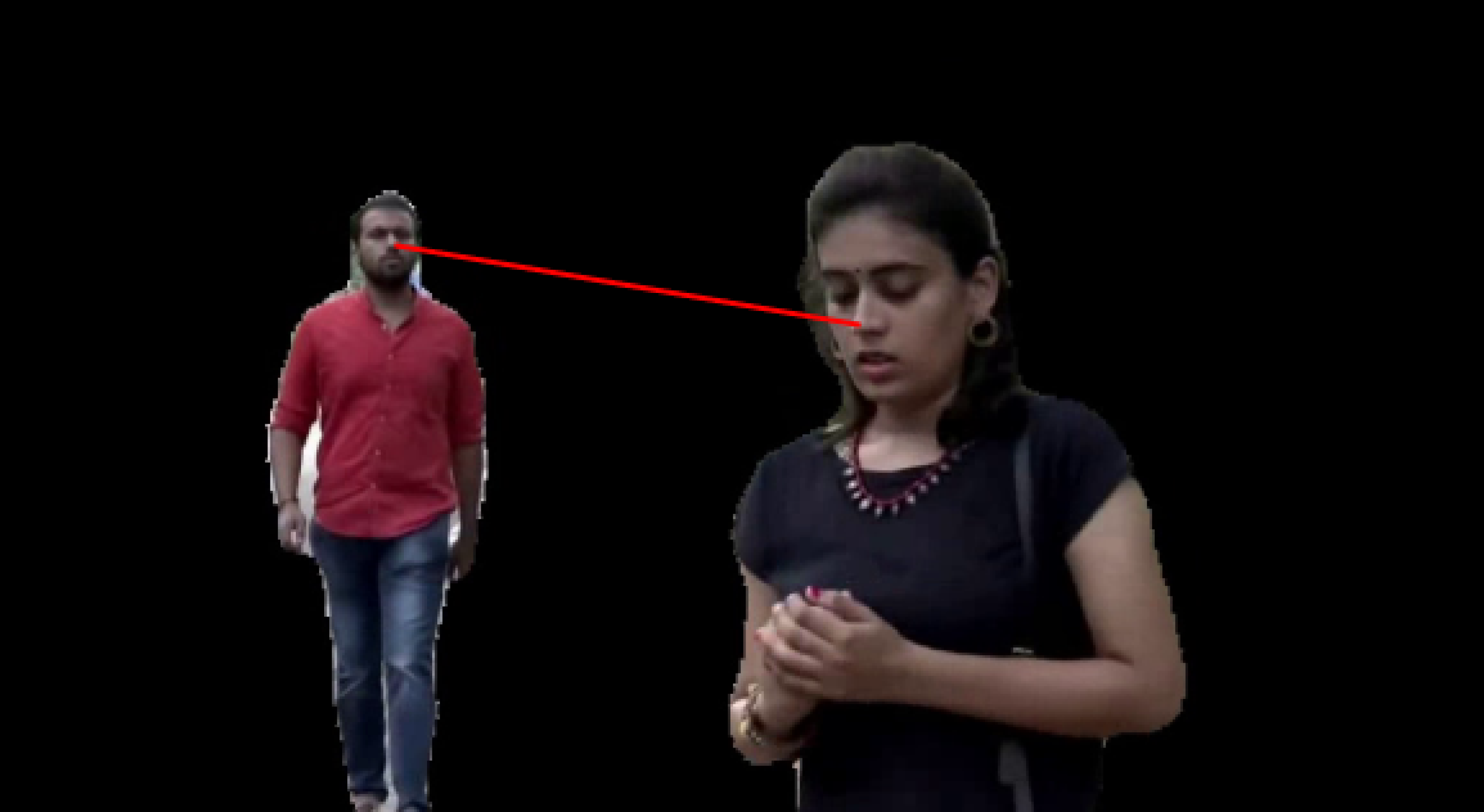}
    \centering(a) Stalking video frame.
  \end{minipage}
    \caption{Comparison of relative distance between non-stalking and stalking cases.
}
    \label{fig5}
\end{figure}

Fig.~\ref{fig5} depicts the relative distance by drawing a line. Following the calculation, a single numerical value for the relative distance from each frame is derived. As a result, we acquired a total of 29 numerical values from various features, including one from relative distance, four from head pose estimation, and twenty-four from facial landmarks. Moreover, it is important to highlight that we are only concerned with two people in order to gain these features, one as a victim and the other as a stalker. If there are numerous victims or stalkers in the video frames, only the front two people's features are retrieved. There are no videos in our dataset with several victims or stalkers in the foreground. In most circumstances, only one victim and one stalker are present, but in certain cases, several stalkers are present where the single victim is on the front side so that no incorrect results are produced.

\label{sec:sample3}
\subsection{DIMENSION TRANSFORMATION}
Initially, the numerical data and extracted frames were not in the right dimension to be put as input into our model. Both data were structured as individual frames. Therefore, it was necessary to convert both sets of data into the form of video. In order to achieve this, the numerical numbers are arranged in a 1 × 29 dimension, representing the compilation of values from each frame. Therefore, the overall dimension for a total of 1190 frames becomes 1190 × 29. The numerical values are then transformed into a dimensional form of  238 × 5 × 29 (representing the number of videos, number of frames, and number of columns) in order to be fed into our model. Likewise, the frames that were initially extracted are resized to dimensions of 128 × 128 × 3. Subsequently, the 1190 frames require a transformation resulting in a new representation of dimensions 238 × 5 × 128 × 128 × 3 (number of videos, number of frames, width, height, number of channels) to be fed into the model. During the process of transformation, careful consideration is given to maintaining the synchronization between frames and numerical values, ensuring that each frame aligns accurately with its associated numerical value. A sorting technique is employed based on the frame name in order to achieve synchronization.

\label{sec:sample3}
\subsection{DATA NORMALIZATION}
In the field of machine learning, the process of data normalization becomes necessary when the features within a dataset demonstrate varying ranges of values. In the absence of normalization, machine learning models may encounter interruptions, require more computational time, and produce incorrect predictions. The motivation behind normalizing the data is to standardize the numbers onto a common scale and ensure uniform mapping. In our dataset, numerical values obtained from facial landmarks and head pose estimation have varying ranges of values. For instance, the distance feature ranges from 91.4 to 1337.78, while the pitch angle feature ranges from -86 to 82, and so on. The range of values in total distributions spans from -86 to 1836. Thus, in order to properly feed the numerical values into the model, normalization is required. In this study, the StandardScaler function from the Python scikit-learn module is utilized to normalize the data.

The Standard Scaler scales the data in such a way that the resulting distribution has a mean value of 0 and a standard deviation of 1. The Standard Scaler function calculates the standard deviation and mean for each feature, then scales each value by subtracting the mean from the value and dividing it by the standard deviation. The data is normalized by applying the following equation:

\begin{equation}
    z = \frac {x - {\mu}} {\sigma}
    \label{p8}
\end{equation}

where, $z$ represents the resulting value, $x$ represents the original value, $\mu$ denotes the mean, and $\sigma$ represents the standard deviation.

\subsection{MODEL SPECIFICATION}
\begin{figure*}
  \centering
  \begin{tabular}{c c c c}
  \includegraphics[width=7.16in]{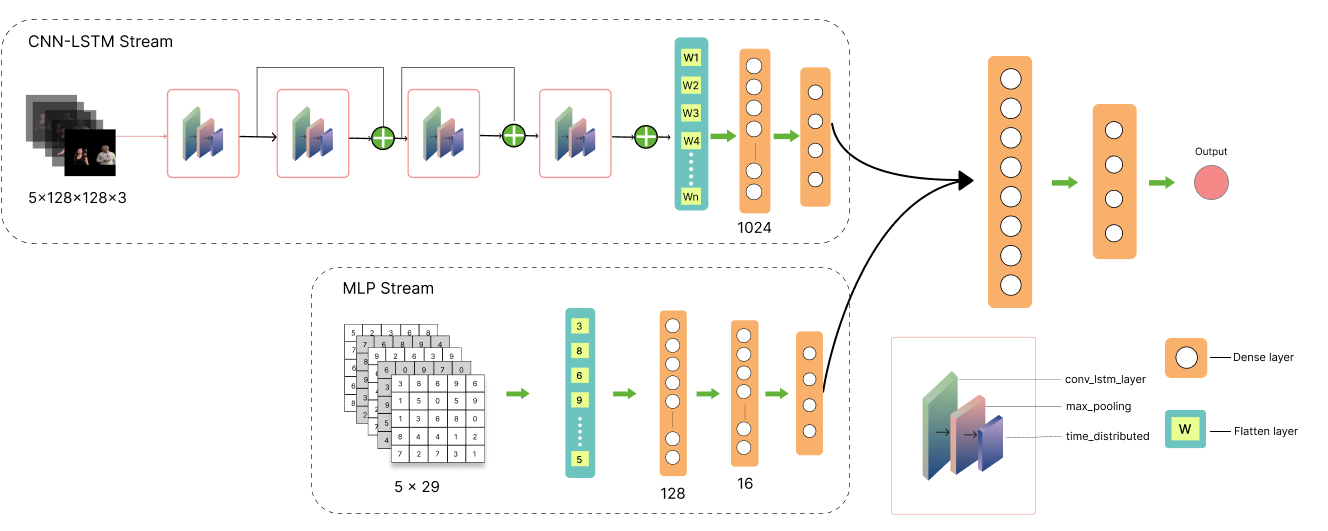} \\
  \end{tabular}
  \caption{Proposed CNN-LSTM-MLP hybrid fusion architecture.
}
    \label{fig6}
\end{figure*}

In this section, we provide a brief description of the classification framework involved in our proposed model, which is shown in Fig.~\ref{fig6}. Our proposed framework consists of two different streams, namely the CNN-LSTM stream and the multilayer perceptron (MLP) stream. The CNN-LSTM stream is implemented to extract spatio-temporal features from image frames. The aim of the MLP stream is to extract features from the numerical data. As an input to the MLP stream, structured numerical data is obtained from facial landmarks where the head pose estimation, and relative distance are provided. In addition, retrieved background subtracted image frames from videos are fed into the CNN-LSTM stream to extract dynamic features. Lately, this model has concatenated the outputs of both streams in a feature fusion process to identify whether the video is stalking or not.

\label{sec:sample3}
\subsubsection{CNN-LSTM-BASED FEATURE EXTRACTION}
In this paper, we decide to apply the popular CNN-LSTM model named ConvLSTM \cite{r27} to extract features from video image frames. ConvLSTM model is proved to be a useful model as a feature extractor in the field of human action recognition \cite{r28}.

Convolutional LSTM is basically an extension of FC-LSTM \cite{r27}. FC-LSTM shows too much redundancy for spatial data. To resolve these issues, Convolutional LSTM has been initiated. Both the input-to-state and the state-to-state transitions in a Convolutional LSTM have a structure that is similar to that of a convolution. By layering numerous Convolutional LSTM layers and building an encoding-forecasting structure, we construct a network model not just for the precipitation nowcasting problem but for other general spatiotemporal sequence forecasting issues like action recognition. When dealing with spatiotemporal data, FC-LSTM suffers from the use of complete connections in transitions between inputs and states when no spatial information is recorded. As a solution to this problem, a distinct aspect of our design is that the Convolutional LSTM’s inputs $X_{1},..., X_{t}$, cell outputs $C_{1},...,C_{t}$, hidden states $H_{1},..., H_{t}$, and gates $i_t$, $f_t$, $o_t$ are all three-dimensional tensors whose final two dimensions are spatial (rows and columns). In the convolutional LSTM structure, it first transforms a 2D image into a 3D tensor. Before performing the convolution, it is necessary to provide padding in order to guarantee that the outputs have the same number of rows and columns as the inputs. By analyzing the inputs and past states of its neighbors, the convolutional LSTM predicts a cell’s future state.

This ConvLSTM architecture is divided into some layers which are given below:
\begin{itemize}
  \item In the ConvLSTM layer, the matrix multiplication step that is normally performed at each gate in the LSTM cell is replaced with a convolution operation.
  \item MaxPooling layers are being used to decrease the size of the data in the images being processed. Basically, max pooling is a pooling technique for determining the maximum value of a feature map’s patches. Additionally, it leads to a faster convergence rate by selecting better invariant features, consequently improving generalization performance.

  \item The Time Distributed layer provides support for the manipulation of data in the form of time series or individual frames extracted from a frame. Because of this, it is feasible to make use of a layer for each of the inputs.

  \item The matrix can be easily converted into an output vector with the help of a flatten layer. The process of a flatten layer usually includes converting each of the dimensional arrays that are created as a consequence of pooling feature mappings into a single extended continuous linear vector.
  
  \item Neurons that are connected to each other are found in the fully connected layer which can be found at the very end of the ConvLSTM architecture. In order to modify the dimension of the vectors or to classify the vectores, a dense layer must be employed, and each neuron must be involved in this process.
\end{itemize}

The suggested ConvLSTM that is implemented in this research has an architecture that comprises of 4 blocks, followed by a flattening layer, and then two fully connected layers. Each block is made up of three layers: the ConvLSTM layer, the MaxPooling3D layer, and the TimeDistributed layer. Even though there are the same number of frames in each block, their individual sizes become smaller as additional layers are added, and there are more channels as a result. Images with dimensions of 5 x 128 x 128 x 3 are fed into the model as input (number of frames, width, height, and number of channels). Following that, the first ConvLSTM layer decreases the number of pixels in the input to 126 x 126 (height x width) while simultaneously increasing the number of channels to 4. A MaxPooling3D layer can be found after the first ConvLSTM layer. This layer halves the size of the frames, making them 63 x 63 (height x width) without affecting the total number of channels. The number of frames in the second ConvLSTM layer of the second block remains the same, but the size of the pixels has been decreased to 61 x 61 and the number of channels has been increased from 4 to 8. Following the completion of MaxPooling3D, it drops the pixel count to 31 x 31, much like the initial MaxPooling3D layer. In a way quite similar to the previous blocks, the third and fourth ConvLSTM layers each have a size that is 29 x 29 and 13 x 13, respectively. Following the third and fourth ConvLSTM layers, there are MaxPooling3D levels that have pixel sizes of 15 x 15 and 7 x 7, respectively, the same as the preceding MaxPooling3D layers. After the first four blocks, there is a flatten layer, followed by two fully connected layers, and these are located at the very end of the network. Between these two fully connected layers, a dropout probability of 0.5, which signifies that a random 50\% of the connections between neurons are eliminated per epoch, is implemented in order to prevent overfitting and reduce dependency. The process of flattening involves transforming the data into a one-dimensional array that consists of 3920 neurons so that it can be input into the fully connected layers. Finally, in the first fully connected layer, there are 1024 neurons, and in the second dense layer, there are just 4 neurons. These 4 neurons are the output of this stream; the output is lately used to concatenate with other neurons in order to accomplish the feature fusion. 

\label{sec:sample3}
\subsubsection{MLP-BASED FEATURE EXTRACTION}
The multilayer perceptron (MLP) is an artificial neural network that operates in a feed-forward manner \cite{r29}. It is composed of an input layer, an output layer, and one or more hidden layers. The input layer is responsible for receiving the input data, which is commonly represented as a vector of features. Every individual neuron within the input layer is representative of a distinct feature, with its assigned value indicating the degree of strength or significance related to that feature. The hidden layers of a multilayer perceptron (MLP) are the components of the network where the process of feature extraction from the input data takes place. In a hidden layer, every neuron receives the outputs of the neurons in previous layers as input, undergoes a linear transformation, and subsequently subjects the outcome to a non-linear activation function. The application of a non-linear activation function enables the neural network to acquire a comprehensive understanding of the complex relationships existing between the input and output data.

In this study, the structured numerical values extracted from the facial landmarks, head pose estimation, and relative distance are fed as inputs to MLP. Since the numerical values are shaped in 3-dimensional order, a flatten layer is constructed after the input layer to convert the data into a 1-dimensional array for feeding the next layer. After the flatten layer, three hidden layers are added in sequential order, where the last hidden layer includes only 4 neurons. The activation function Rectified Linear Unit (ReLu) is used in this MLP stream. The ReLu activation function removes all negative values and maps positive ones directly. This activation function is selected because of its computational efficiency and because it is inexpensive to use. The output of the last hidden layer lately concatenates with the output of the CNN-LSTM stream. The proposed MLP design is shown in Fig.~\ref{fig6}.

\label{sec:sample3}
\subsubsection{FEATURE FUSION}
The implementation of feature fusion in this study follows a densely connected multi-layer perceptron (MLP) approach. The concatenation of the output from the CNN-LSTM stream and the MLP stream, each consisting of 4 neurons, occurs at the initial dense layer of feature fusion. The fusion information passes through two fully connected dense layers, each utilizing the Rectified Linear Unit (ReLU) activation function. After that, the following layer consists of an additional dense layer, which acts as the final layer, incorporating a sigmoid activation function.  The final layer of this model consists just of one neuron, which is utilized for the purpose of predicting the classification. Finally, the output layer is responsible for generating the predicted classification result.

\subsection{THE PROPOSED CNN-LSTM-MLP HYBRID FUSION MODEL}
\label{sec:sample2}
The algorithm begins with the input dataset $V = \{V_1, V_2, \ldots, V_n\}$. In the first step, a frame extraction function is applied to $V$, resulting in a set of image frames $V_f=\{V_{i_1},V_{i_2},\ldots,V_{i_n}\}$. Subsequently, in step 2, a background subtraction function processes $V_f$ to produce a set of segmented image frames $V_{B_s}$, where backgrounds are removed, leaving only the objects. In the third step, facial landmark detection is performed on $V_{B_s}$, yielding a set of prioritized facial points $D_p$, with six selected points for each individual in each frame from 68 facial points. Moreover, for 2 individuals a set of coordinate points $C_p$ is also calculated. Moving to step 4, a head pose estimation function takes $V_{B_s}$ and $D_p$ as inputs, generating numerical values $N_p$ derived from yaw and pitch angles for both individuals. Step 5 involves measuring the relative distance between the two individuals in $V_{B_s}$ using $D_p$, resulting in a single distance value $d$. Subsequently, in step 6, the algorithm employs data normalization on the calculated values from steps 3, 4, and 5 which are $C_p$, $N_p$, and $d$, respectively yielding normalized numerical data $F_n$. In step 7, a dimension transformation method is applied to $F_n$, producing transformed data $D_F$. The transformed data $D_F$ is then fed into an MLP function in step 8, resulting in $M_1$, which consists of four neurons. Following this, in step 9, a resize function processes the segmented image frames $V_{B_s}$ to generate resized frames $R_f$. Another dimension transformation is applied to $R_f$ in step 10, producing $D_{F_1}$. Steps 11 and 12 involve the application of a ConvLSTM function to $D_{F_1}$, generating $S$ with four neurons, and a feature fusion model that takes $S$ and $M_1$ as parameters to produce the final output, which signifies the identification of a stalking or non-stalking scenario.

\begin{algorithm}[h]
 \caption{Proposed CNN-LSTM-MLP Hybrid Fusion Model}
 \begin{algorithmic}
 \STATE  \textbf{Input:} Dataset $V$  contains $n$  Videos:  $V \leftarrow V_1, V_2, \ldots, V_n$ 
 \vspace{-4mm}
 \STATE \textbf{Output:} Predicted class, 1 neuron
    \STATE \textbf{Step 1:} Frame Extraction ($V$): 
    \FOR {each video $V$ }
    \STATE Extract Frame $V_f \leftarrow V_{i_1}, \ldots, V_{i_n}$ 
    \ENDFOR  
    \STATE \textbf{Step 2:} Background Subtraction ($V_f$): 
    \FOR {each frame in $V_f$ }
    \STATE Segmented Frames $V_{B_s} \leftarrow$ Perform\_Back\_Subtraction 
    \ENDFOR
    \STATE \textbf{Step 3:} Facial Landmark ($V_{B_s}$): 
    \STATE $D \leftarrow$ All 68 landmarks from the model 
    \STATE Prioritized\_points, $D_p \leftarrow$ 6 points from 68 
    \STATE two person's pixel\_points, $P_p \leftarrow D_p \times 2$
    \STATE two person's co\_ordinate\_points, $C_p \leftarrow P_p \times 2$ 
    \STATE  \textbf{Step 4:} Head Pose ($V_{B_s} , D_p $): 
     \STATE $\theta_1 \leftarrow$ Calculated yaw\_angle 
     \STATE $\theta_2 \leftarrow$ Calculated pitch\_angle 
     \STATE two person's calculated 4 numerical points
    \STATE From $\theta_1,\theta_2 \rightarrow N_p$
    \STATE \textbf{Step 5:} Distance Measurement ($V_{B_s} , D_p $): 
    \STATE $N_{T_1} \leftarrow (V_x, V_y)$ 
    \STATE $N_{T_2} \leftarrow (S_x, S_y)$
    \[ d = \sqrt {({ V_{x} - S_{x} })^2  + ({ V_{y} - S_{y} })^2} \]
    \STATE \textbf{Step 6:} Data Normalization ($C_p, N_p, d $): 
    \[
    z = \frac {x - \mu} {\sigma}
    \]
   \STATE Final Numerical data $\rightarrow F_N$
    \STATE \textbf{Step 7:} Dimension Transformation ($F_N$): 
     \STATE $F_N$ transformed into dimensional form $\rightarrow D_F$
    \STATE \textbf{Step 8:} MLP ($D_F$ ):
    \STATE $M_1 \leftarrow$ output of last hidden layer, 4 neurons
    \STATE \textbf{Step 9:} Resize ($V_{B_s}$):
    \STATE $R_F \leftarrow$ Resized Frames from $V_{B_s}$
    \STATE \textbf{Step 10:} Dimension Transformation ($R_F$):
    \STATE $R_F$ transformed into dimensional form $\rightarrow D_{F_1}$
    \STATE \textbf{Step 11:} ConvLSTM ($D_{F_1}$):
    \STATE output\_sequence, $S =$ convlstm-layer $(D_{F_1})$
    \STATE $S \leftarrow 4$ neurons
    \STATE \textbf{Step 12:} Feature Fusion ($S , M_1$):
    \STATE Output $\leftarrow$ Concatenation of neurons from $S$ and $M_1$, resulting in 1 neuron

\end{algorithmic}
\end{algorithm}

\section{EXPERIMENTS AND RESULTS}
\label{EXPERIMENTS AND RESULTS}
\subsection{IMPLEMENTATION DETAILS}
Our suggested approach is implemented in two phases due to limitations in resources and computational power. The first phase implements frame extraction, background subtraction, face landmarks, angles, and distance measurement using Python 3.7.4 with Pycharm 2021 2.2 as the development environment. This process is executed in the Windows 10 64-bit operating system, utilizing an Intel (R) Core i5-8400 processor and 16GB of RAM. The final phase includes the implementation of dimension transformation, data normalization, and the hybrid fusion model in the Google Colab notebook environment. Google Colab offers free access to a GPU, namely the Tesla T4 GPU. Additionally, it provides pre-installed Python 3.x packages and the Keras API, which is facilitated by the Tensor-Flow backend.

In the experiment, the dataset is divided into three subsets: training, validation, and testing. This division followed a split of 60\%, 20\%, and 20\% accordingly, where each subset consists of video frames and their corresponding numerical values. In regard to parameter settings, all models, including our proposed model and other models for comparison, are set to the same hyperparameters. The optimizer has been specified as Adam with a learning rate of $1*10^{-3}$ and a decay of $5*10^{-4}$. Early stopping callback is defined to stop training when there is no improvement in validation loss in order to prevent overfitting. The early stopping callback is responsible for determining the optimal number of epochs required for the model to achieve its highest accuracy during the iteration process. The proposed model achieves the highest level of accuracy in both the training and validation sets after 35 epochs. It was noticed that when the model was run for additional epochs, it began to overfit.

\subsection{PERFORMANCE EVALUATION}
The experimental results of our proposed model are presented here. Firstly, we show the comparison between with and without fusion of our proposed method. Secondly, we compare the proposed method with three other widely popular methods. The first is VGG16 \cite{r30} , and it consists of 13 convolutional layers, 5 pooling layers, and 3 fully connected layers. For the sake of comparison, this architecture is defined as only a CNN stream. Following that, we compare ConvLSTM without Fusion, which consists of a CNN-LSTM stream. Lastly, a comparison is made with the CNN-MLP stream, which combines VGG16 as a CNN stream and an MLP stream with three other fully connected layers, in which fusion is incorporated. It is important to note that fusion includes the numerical values obtained from facial features, but without fusion, such values are excluded. Finally, we compare our proposed method with other state-of-the-art methods, showing that our proposed method outperforms other methods.

To measure the robustness and efficiency of any model, some parameters are needed, which include accuracy, precision, recall, and F-measure. Accuracy refers to the ratio of the number of correctly predicted videos to the total number of videos. Precision refers to the ratio of the number of true positives (TP) to the total number of true and false positives (FP). Recall refers to the proportion of true positives relative to the total number of true positives as well as false negatives (FN). The F-measure is a normalized harmonic mean that takes into account both precision and recall. The better the projected accuracy of the algorithm is going to be, the nearest the value of the F-measure is to being equal to 1.0. These parameter-specific calculations are defined as follows:

\begin{equation}
  Accuracy \ =  \ \frac { TP+TN } {TP+FP+TN+FN}
  \label{p9}
\end{equation}

\begin{equation}
  Precision \ = \  \frac { TP } {TP+FP}
  \label{p10}
\end{equation}

\begin{equation}
  Recall \ = \  \frac { TP } {TP+FN}
  \label{p11}
\end{equation}

\begin{equation}
  F-measure \ = \  2 * \frac { Precision * Recall } { Precision + Recall }
  \label{p12}
\end{equation}

In the above equations [\ref{p9}, \ref{p10}, \ref{p11}, \ref{p12}], TP (true positive) represents the number of correctly predicted non-stalking videos. FP (false positive) represents the number of incorrectly predicted non-stalking videos. TN (true negative) represents the number of correctly predicted stalking videos. FN (false negative) represents the number of incorrectly predicted stalking videos.

\begin{table}[h]
\centering
\caption{Performance analysis of with and without fusion}
\label{t1}
\begin{tabular}{@{}llll@{}}
\\
\toprule
Model            & \ \ \ \   Metrics    & \  Non-stalking    & \ Stalking    \\
\midrule
\\
\begin{tabular}[c]{@{}c@{}}Proposed \\ (without fusion)\end{tabular}    & \ \ \  \begin{tabular}[c]{@{}c@{}c@{}}Precision \\ \\  Recall \\ \\ F-measure \end{tabular} &  \begin{tabular}[c]{@{}c@{}c@{}}\ \ \ \ \ \  0.71 \\ \\  \ \ \ \ \ \ 0.71\\ \\  \ \ \ \ \ \ 0.71 \end{tabular} &  \begin{tabular}[c]{@{}c@{}c@{}} \ \ \ 0.78 \\ \\  \ \ \ 0.78 \\ \\ \ \ \ 0.78 \end{tabular}  \\  \\  \midrule
\\
\begin{tabular}[c]{@{}c@{}}Proposed \\ (with fusion)\end{tabular}    & \ \ \  \begin{tabular}[c]{@{}c@{}c@{}}Precision \\ \\  Recall \\ \\ F-measure \end{tabular} &  \begin{tabular}[c]{@{}c@{}c@{}}\ \ \ \ \ \  0.83 \\ \\  \ \ \ \ \ \ \textbf{0.95}\\ \\  \ \ \ \ \ \ 0.89 \end{tabular} &  \begin{tabular}[c]{@{}c@{}c@{}} \ \ \ \textbf{0.96} \\ \\  \ \ \ 0.85 \\ \\ \ \ \ \textbf{0.90} \end{tabular}  \\ \\ \bottomrule

\end{tabular}
\end{table}


In Table~\ref{t1}, a comparative analysis between the proposed architecture with and without fusion is shown. The table shows that with fusion, the model's precision in the stalking class is 0.96, but without fusion, it is 0.78. Furthermore, the precision in the non-stalking class is 0.83 with fusion and 0.71 without fusion, which is significantly lower. The recall values with fusion are 0.95 and 0.85 for the non-stalking and stalking classes, respectively, whereas the values without fusion are 0.71 and 0.78.  Furthermore, when fusion is not included, other evaluation metrics, such as F-measure scores, are lower. The results demonstrate that the proposed approach with fusion achieves better results than the method without fusion, meaning that the numerical values obtained from the facial features have a major impact on achieving better results.

\begin{figure}[htp]
    \centering
    \includegraphics[width=3.5in]{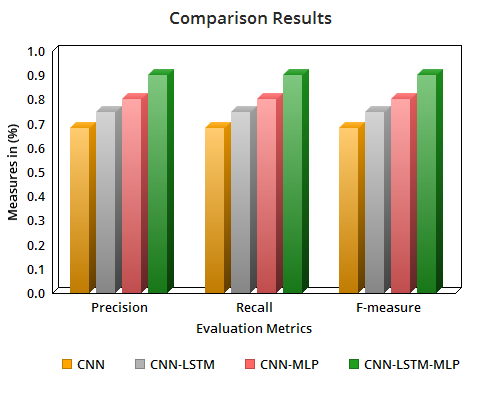}
    \caption{Comparison results with different methods.}
    \label{fig7}
\end{figure}

The comparative results from four different types of approaches are presented in Fig.~\ref{fig7}. The CNN-LSTM-MLP (proposed) and CNN-MLP include fusion, whereas the CNN-LSTM and CNN do not include fusion. In CNN and CNN-MLP architectures, the widely popular VGG16 architecture is used as a CNN stream to extract features from frames. The graphical representation of macro-averaged precision, recall, and f-measure of these four methods is shown in Fig.~\ref{fig7}. CNN provides the lowest precision of 0.68. Furthermore, the recall and f-measure scores are also 0.68. CNN-LSTM performs slightly better, with 0.75 in all metrics. CNN-MLP surpasses CNN-LSTM in terms of performance. CNN-MLP produces a precision of 0.80, a recall of 0.80, and an F1 score of 0.80. When compared to these three methods, CNN-LSTM-MLP provides the greatest results, with a precision of 0.90 and the same results in other metrics.

The training, validation, and testing accuracies of four different models are presented in Table~\ref{t2}. It is clear that when facial features are used as numerical values in the fusion technique, results improve dramatically. The inclusion of fusion in CNN-MLP improves training accuracy to 80.53\%, while CNN alone acquires 68.63\% accuracy in training; the same applies to both validation and training accuracy. Also, the CNN-LSTM-MLP architecture, which incorporates fusion, has higher training, validation, and testing accuracies compared to the CNN-LSTM architecture. Furthermore, it is necessary to highlight that the CNN and CNN-MLP models alone classify individual frames, but the CNN-LSTM and CNN-LSTM-MLP models incorporate LSTM to classify a sequence of frames, which forms a video. A stalking scene cannot be predicted by a single frame since stalking requires a minimum time span. Therefore, the CNN-LSTM-MLP model is chosen as it effectively classifies a whole video and provides better results.

\begin{table}[h]
\centering
\caption{Performance Evaluation of the four models.}
\label{t2}
\begin{tabular}{@{}llllll@{}}
\\
\toprule
   CNN                & LSTM    & MLP    & \begin{tabular}[c]{@{}c@{}}Training \\ accuracy\end{tabular}  & \begin{tabular}[c]{@{}c@{}}Validation \\ accuracy\end{tabular} & \begin{tabular}[c]{@{}c@{}}Testing \\ accuracy\end{tabular}  \\ \midrule \\
   \ \ \checkmark &                  &                  & \ 68.63\% & \ \ 68.49\% & \ 68.00\% \\ \\
   \ \ \checkmark & \ \ \ \checkmark &                  & \ 74.65\% & \ \ 72.92\% & \ 75.00\% \\ \\
   \ \ \checkmark &                  & \ \  \checkmark  & \ 80.53\% & \ \ 73.95\% & \ 79.83\% \\ \\
   \ \ \checkmark & \ \ \ \checkmark & \ \  \checkmark  & \ \textbf{93.93\%} & \ \ \textbf{90.26\%} & \ \textbf{89.58\%} \\ \\
    
    \bottomrule
\end{tabular}
\end{table}

\begin{figure}[!tbp]
  \centering
  \begin{minipage}[b]{0.5\textwidth}
    \includegraphics[width=3.5in]{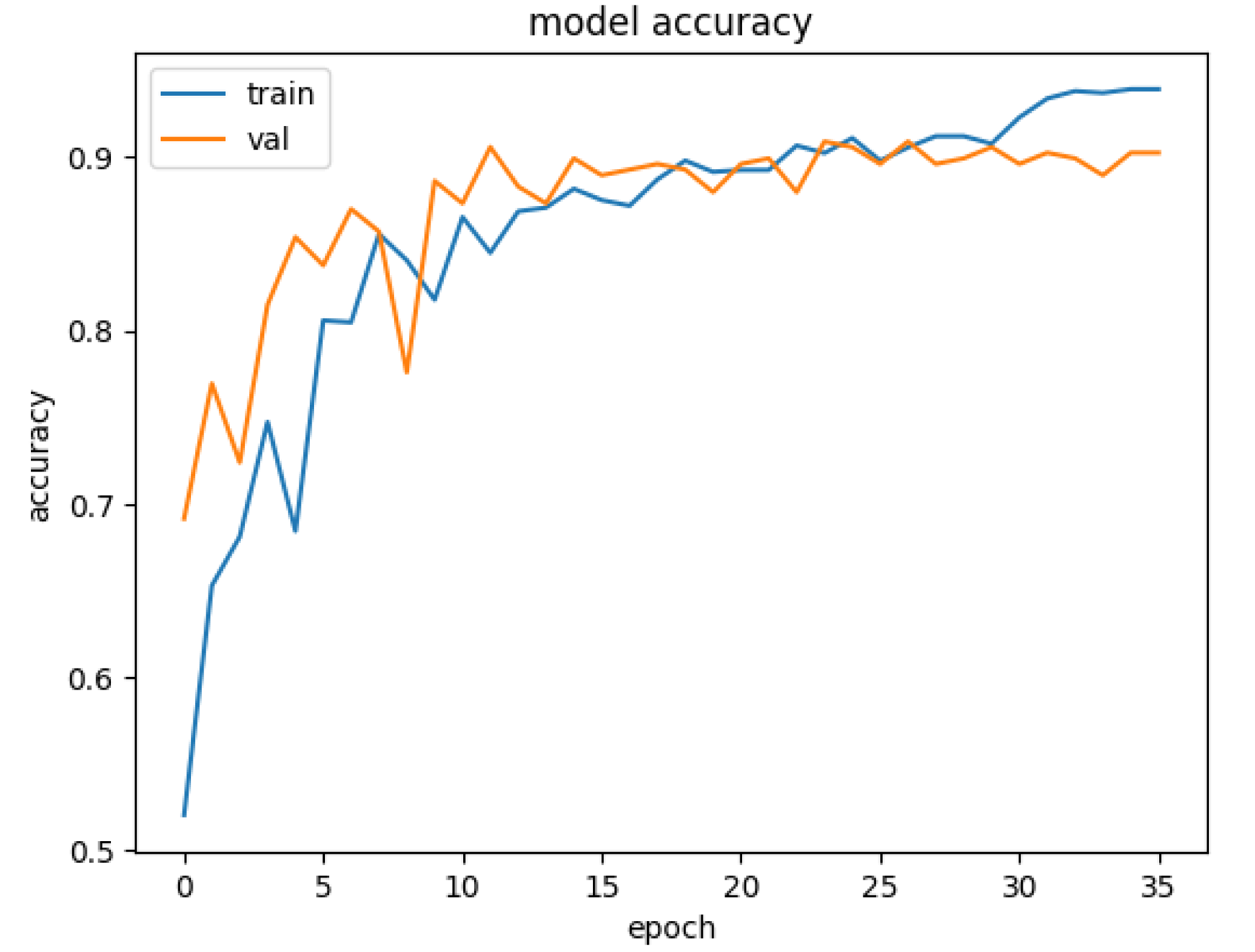}
\centering(a) Training and validation accuracy.
  \end{minipage}

  \hfill
  \begin{minipage}[b]{0.5\textwidth}
    \includegraphics[width=3.5in]{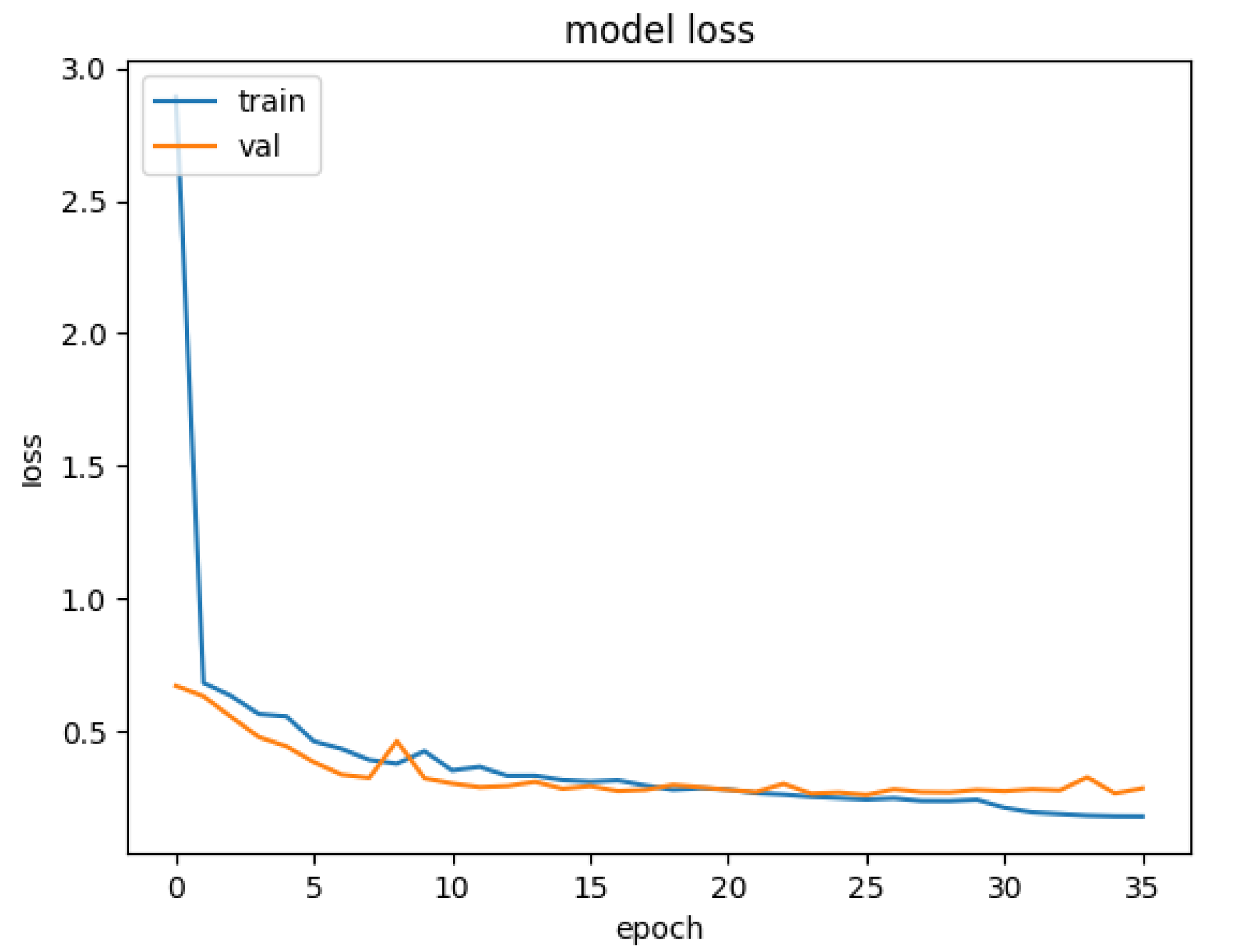}
\centering (b) Training and validation loss.
  \end{minipage}
    \caption{ Accuracy and loss during training and validation of the proposed model.}
    \label{fig8}
\end{figure}

\begin{figure}[htp]
    \centering
    \includegraphics[width=3.5in]{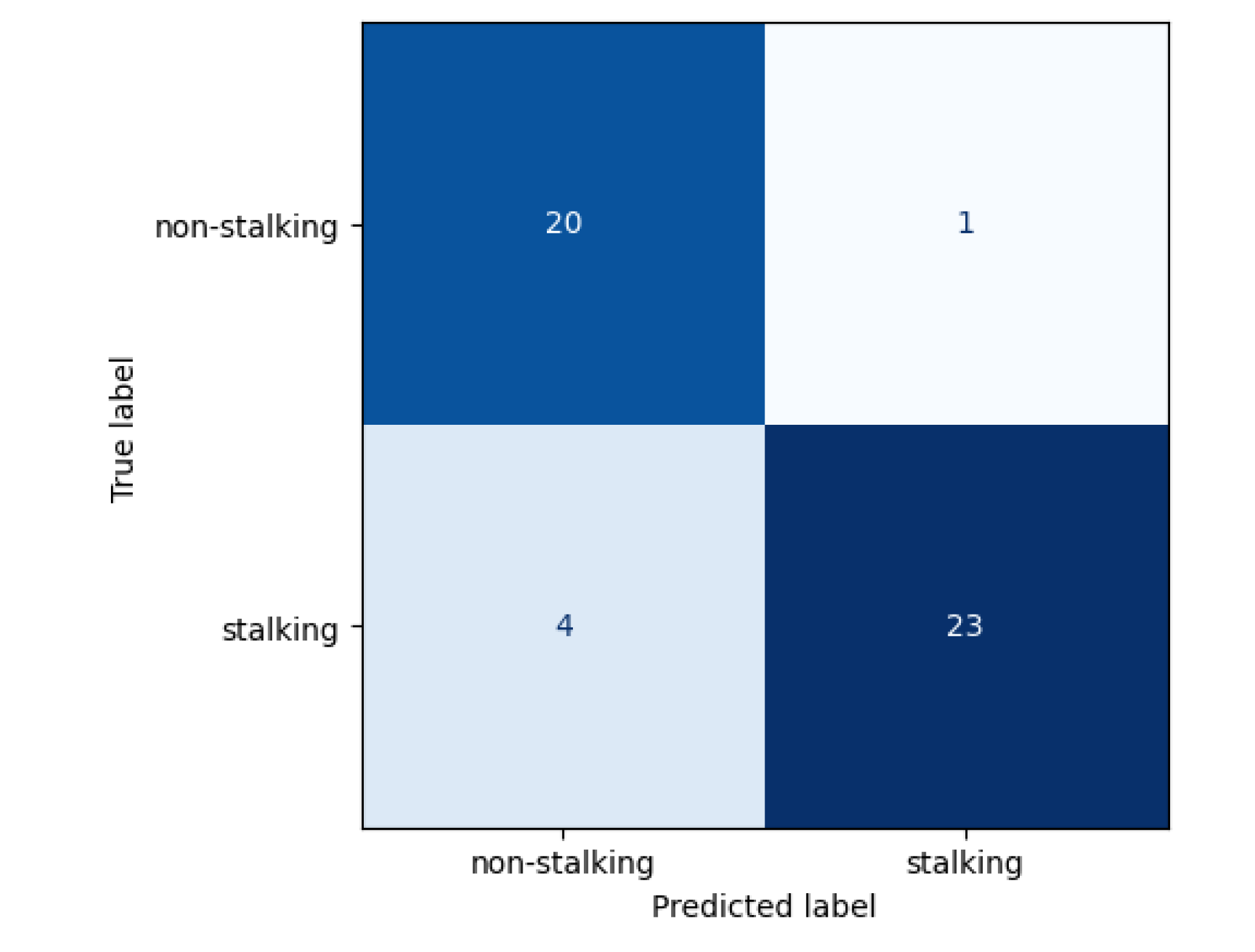}
    \caption{Confusion matrix of the proposed model.}
    \label{fig9}
\end{figure}

Our proposed model is trained for a total of 35 epochs in order to achieve the greatest accuracy and extract its full potential. The model reaches a validation accuracy of 90.26\% and a training accuracy of 93.93\% following the training process. The accuracy and loss of the proposed model are illustrated in Fig.~\ref{fig8}. In addition, the proposed model's confusion matrix is shown in Fig.~\ref{fig9}. Out of 48 samples, the overall number of correct predictions is 43, with 20 correctly identified non-stalking videos and 23 correctly classified stalking videos. The total number of misclassified videos is 5, with four from the stalking class and one from the non-stalking class. The results show that our model performs well in detecting both cases.

We also compare our method with other state-of-the-art methods proposed in previous studies. There have been many studies on street violence, but very few on stalking. For comparison with our findings, a table including the most relevant data from comparable prior work is provided. The comparison results are shown in Table~\ref{t3}. 

\begin{table}[!htp]
\centering
\caption{Comparison between the suggested approach and existing state-of-the-art approaches.}
\label{t3}
\begin{tabular}{@{}lllll@{}}
\\
\toprule
\\
Literature            &   Year    & \ \ \ \ \ \ \ \ Dataset    & \ Method used  & Accuracy    \\
\\
\midrule
\\
\begin{tabular}[c]{@{}c@{}}Liu et \\ al. {}\cite{r13}{}\end{tabular}    & 2019 & \ \ \begin{tabular}[c]{@{}c@{}c@{}}Multi-appearance-  \\  based Custom \\ Dataset\end{tabular} & \!\!\!\!\!\! \begin{tabular}[c]{@{}c@{}}Neoface + Stalker \\ Query\end{tabular} & \!\!\!\!\!\! \begin{tabular}[c]{@{}c@{}}100\% \\ (in lab level)\end{tabular} \\
\\
\\
\textbf{Proposed}    & \textbf{2023} & \begin{tabular}[c]{@{}c@{}c@{}}\textbf{Single-appearance-}  \\  \textbf{based Custom} \\ \textbf{Dataset}\end{tabular} & \ \begin{tabular}[c]{@{}c@{}c@{}}\textbf{CNN-LSTM-}  \\  \textbf{MLP-based} \\ \textbf{Fusion}\end{tabular} & \ \ \textbf{89.58\%} \\ \\ \bottomrule
\end{tabular}
\end{table}

One important point to note is that many researchers have worked on methods for identifying anomalous human behavior in various scenarios. Their primary objective was to identify actions from their selected dataset, which did not include stalking. Their approach can recognize deviant behaviors such as loitering, fighting, kicking, and stealing. A special mention to Liu et al.\cite{r13} for initiating significant work on stalker detection and opening the doors for future researchers. They claim that theirs is the first devoted investigation into stalker detection and that it has been a huge success. However, the authors of the study acknowledge certain limitations, which they outline in the discussions and open challenges section. They do not disclose their real accuracy but claim that accuracy in the laboratory is 100\%. Additionally, their dataset contains videos collected from multiple cameras that are not publicly available. Besides, the primary difference between their work and ours is that they utilize various videos from several situations to identify stalking scenarios, which can be defined as multiple appearances. Due to this, they need person re-identification. Whereas we just use frames from a single appearance-based scenario, making our work more unique and uncomplicated in detecting stalking scenarios. In our study, we not only examine the facial landmarks of the individuals but also their head pose angles and relative distance, which greatly facilitate our fusion model and achieved an accuracy of 89.58\%. Despite the use of several features and the utilization of a computer vision approach to address such a complicated real-life problem, the result is significant.

\section{CONCLUSION}
\label{CONCLUSION}
Wrapping up the paper, if we examine it closely, we can see that before any form of crime occurs, the criminal will have a plan, and stalking is a common and must-do item to happen first. Because this is often one of the first phases of a crime, if it can be avoided, a large sum of a misdemeanor, such as lady badgering, on-road burglary, and so on, can be reduced. In this work, we use a fusion model that combines CNN, LSTM, and MLP in an attempt to take into consideration multiple facial features to produce more dynamic results than earlier research. However, due to the scarcity of data, more studies are not being conducted to address this real-world issue. We attempted to collect surveillance videos at first but faced some difficulties because privacy is a major problem in this area. In the future, we would like to investigate the effectiveness of stalker detection by implementing such a model in a surveillance system. Furthermore, stalking detection requires more attention so that many researchers can contribute to this topic. We believe we are the second contributor to detecting such complicated action. We anticipate that other researchers will come to contribute to this problem being influenced by this work.

\begin{IEEEbiography}[{\includegraphics[width=1in,height=1.25in,clip,keepaspectratio]{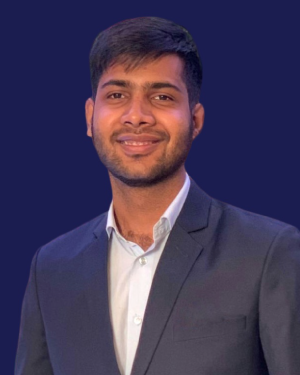}}]{MURAD HASAN } received his B.Sc. in computer science and engineering from BRAC University, Dhaka, Bangladesh, in 2022. During his undergrad, he worked as an Undergraduate Teaching Assistant, which allowed him to assist students during lab and consultation hours. After his undergrad, he worked as a Software Engineer Intern at Samsung R\&D Institute Bangladesh. Afterward, he joined as a Contractual Lecturer at BRAC University, Dhaka, Bangladesh. In addition, he has been honored with the distinction of receiving the Vice Chancellor's Award, being named to the Dean's List, and being awarded a scholarship based on merit.

His research interests include computer vision, machine learning, and image processing. He has a wish not to limit his research only to these areas, since he has a multidisciplinary curiosity including sports vision, AI ethics, cognitive science, and cognitive psychology.
\end{IEEEbiography}

\begin{IEEEbiography}[{\includegraphics[width=1in,height=1.25in,clip,keepaspectratio]{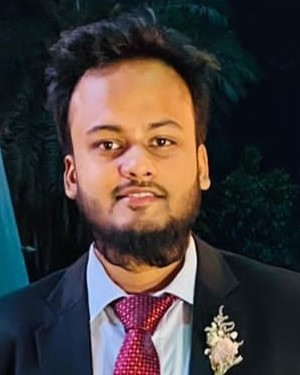}}]{SHAHRIAR IQBAL } received his B.Sc. in computer science and engineering from BRAC University, Dhaka, Bangladesh, in 2022. Throughout his academic career, Shahriar has acquired a significant interest in and an intense passion for cybersecurity, system security, computer vision, and information system management. In addition, he has been named to the Dean's List and VC’s List many times in the period of his undergrad. After his undergrad, he worked as a Software QA Engineer Intern at a1qa, a pure US-based software testing company. Afterward, he joined as a Software QA Engineer at CloudlyIO which is also a US-based company. He presently works as a Software QA Engineer at MyMedicalHub Corp., a firm based in the United States. He actively contributes to the quality and security of software products. His enthusiasm for cybersecurity and commitment to excellence drive his quest for expertise in cutting-edge technology. Moreover, he is aiming to pursue a master’s degree in his desired field of interest.
\end{IEEEbiography}

\begin{IEEEbiography}[{\includegraphics[width=1in,height=1.25in,clip,keepaspectratio]{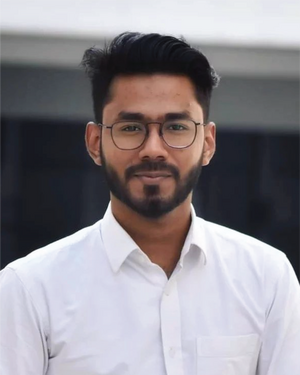}}]{MD. BILLAL HOSSAIN FAISAL } received the B.Sc. degree in computer science from BRAC University, Dhaka, Bangladesh, in 2022. During his time at the university, he showed great interest in software engineering and management information systems. His research interests include artificial intelligence, machine learning, data science, web development, UX, and business intelligence analytics. In his early career, he worked as a UI/UX designer in a Cyber Security company called "Techforing Limited". Currently, he is working as a Project Manager at "Chaldal Limited". Besides that, he is involved as a technical management lead of the Bangladesh Internet Governance Forum.
\end{IEEEbiography}

\begin{IEEEbiography}[{\includegraphics[width=1in,height=1.25in,clip,keepaspectratio]{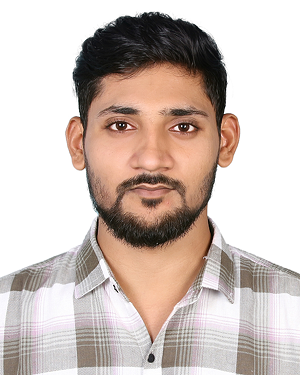}}]{MD. MUSNAD HOSSIN NELOY } received the B.Sc. degree in computer science from BRAC University, Dhaka, Bangladesh, in 2022. He was particularly interested in programming, computer networks, and android application development during his undergraduate years. He is currently seeking a career in government service.
\end{IEEEbiography}

\begin{IEEEbiography}[{\includegraphics[width=1in,height=1.25in,clip,keepaspectratio]{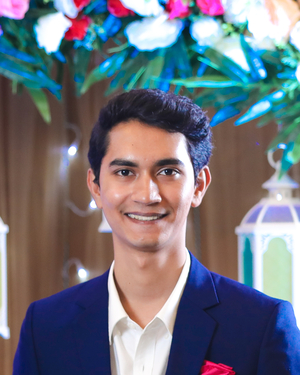}}]{MD. TONMOY KABIR } received the B.Sc. degree in computer science and engineering from BRAC University, Dhaka, Bangladesh, in 2022. During his undergraduate studies, he was particularly interested in software engineering, and management information systems. His research interests include artificial intelligence, machine learning, data science, cyber security, and business intelligence analytics. Presently, he is engaged in the pursuit of his M.Sc. degree in information technology at Griffith University in Brisbane, Queensland, Australia. Additionally, he is an associate member of the Australian Computer Society.
\end{IEEEbiography}

\begin{IEEEbiography}[{\includegraphics[width=1in,height=1.25in,clip,keepaspectratio]{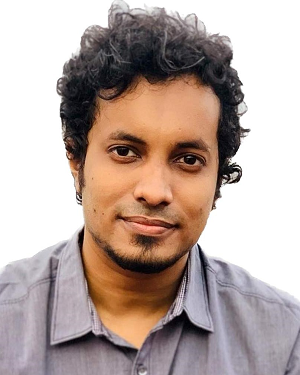}}]{MD. TANZIM REZA } completed his B.Sc. in Computer Science and Engineering from BRAC University in 2018. Afterward, he joined as a contractual lecturer in 2019 and got promoted to full-time lecturer in the Spring of 2020. His research interest is Artificial Intelligence, Machine Learning, Natural Language Processing, and Computer Vision. He has experience in using machine learning algorithms with Python on large-scale datasets, writing scientific research papers, objectoriented analysis design, and game development using java.
\end{IEEEbiography}

\begin{IEEEbiography}[{\includegraphics[width=1in,height=1.25in,clip,keepaspectratio]{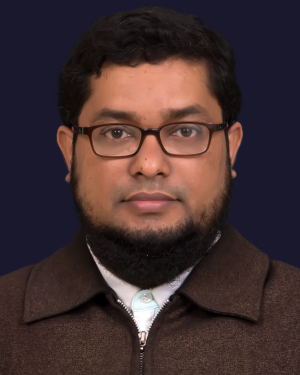}}]{MD. GOLAM RABIUL ALAM } (Member, IEEE) received the B.S. and M.S. degrees in computer science and engineering and information technology, respectively, and the Ph.D. degree in computer engineering from Kyung Hee University, South Korea, in 2017. He also served as a Postdoctoral Researcher with the Computer Science and Engineering Department, Kyung Hee University, from March 2017 to February 2018. He is currently a Professor with the Department of Computer Science and Engineering, BRAC University, Bangladesh. His research interests include health-care informatics, mobile cloud and edge computing, ambient intelligence, and persuasive technology. He is a member of the IEEE IES, CES, CS, SPS, CIS, KIISE, and IEEE ComSoc. He also received several best paper awards in prestigious conferences.
\end{IEEEbiography}

\begin{IEEEbiography}[{\includegraphics[width=1in,height=1.25in,clip,keepaspectratio]{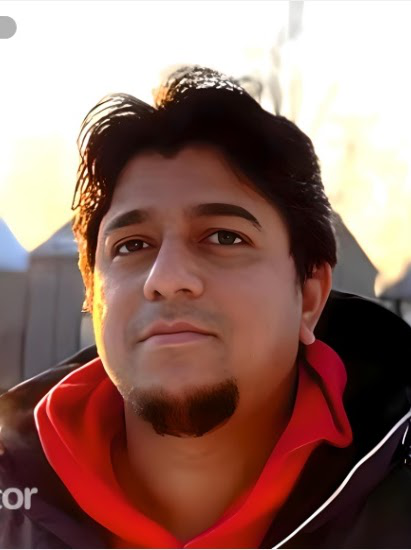}}]{Md Zia Uddin } (Senior Member, IEEE)  completed his Ph.D. degree in Biomedical Engineering in 2011 from Kyung Hee University of South Korea. He is currently working as a Senior Research Scientist at SINTEF Digital, Oslo, Norway. His research is mainly focused on artificial intelligence, machine learning, pattern recognition, sensors,  HCI, assisted living, etc. He has been mostly exploring and applying machine learning methods on data from a wide range of sources to recognize/predict  underlying events such as users’ mood, behaviour, health status, sentiments, etc. He has got more than 150 peer-reviewed research publications (around 75 as a leading author) including book, book chapters, conferences, and journals such as Information Fusion, IEEE transactions on consumer electronics, IEEE Sensors, ACM Transactions on Multimedia Computing, Communications, and Applications, etc. His google scholar citations are more than 4000. Dr. Zia has been working/leading in different work packages of national and international research projects. He has been  enlisted in the World’s Top 2\% Scientists (both career-long \& single year-based), prepared by the Stanford University of USA and Elsevier BV.
 \end{IEEEbiography}

\end{document}